%% file: main.tex
\documentclass{article}

\PassOptionsToPackage{numbers,sort&compress}{natbib}

\usepackage[preprint]{neurips_2026}

\usepackage[utf8]{inputenc} 
\usepackage[T1]{fontenc}    
\usepackage[pagebackref]{hyperref}       
\usepackage{url}            
\usepackage{booktabs}       
\usepackage{amsfonts}       
\usepackage{nicefrac}       
\usepackage{microtype}      
\usepackage[table,dvipsnames]{xcolor}         
\usepackage{graphicx}
\usepackage{multirow}
\usepackage{tabularx}
\usepackage{subcaption}

\title{A General Concept-based Decomposition\\for Vision-Language Embeddings}

\author{%
  Simone Alberto Peirone\thanks{Work carried out while visiting EPFL.} \\
  Politecnico di Torino\\
  Torino, Italy\\
  \texttt{simone.peirone@polito.it} \\
  \And
  Ortal Senouf \\
  EPFL \\
  Lausanne, Switzerland\\
  \texttt{ortal.senouf@epfl.ch} \\
  \And
  Francesca Pistilli \\
  Politecnico di Torino\\
  Torino, Italy\\
  \texttt{francesca.pistilli@polito.it} \\
  \And
  Giuseppe Averta \\
  Politecnico di Torino\\
  Torino, Italy\\
  \texttt{giuseppe.averta@polito.it} \\
  \And
  Pascal Frossard \\
  EPFL\\
  Lausanne, Switzerland\\
  \texttt{pascal.frossard@epfl.ch} \\
}
\input{preamble}

\begin{document}

\maketitle

\input{sections/abstract}

\input{sections/introduction}
\input{sections/related_works}
\input{sections/method}
\input{sections/experiments}
\input{sections/conclusions.tex}

\section*{Acknowledgments}
This study was carried out within the FAIR - Future Artificial Intelligence Research and received funding from the European Union Next-GenerationEU (PIANO NAZIONALE DI RIPRESA E RESILIENZA (PNRR) – MISSIONE 4 COMPONENTE 2, INVESTIMENTO 1.3 – D.D. 1555 11/10/2022, PE00000013). This manuscript reflects only the authors’ views and opinions, neither the European Union nor the European Commission can be considered responsible for them. We acknowledge the CINECA award under the ISCRA initiative, for the availability of high-performance computing resources and support. 
We also acknowledge the support of the European Laboratory for Learning and Intelligent Systems (ELLIS). Simone's work was carried out during a research visit supported by the ``ELSA – European Lighthouse on Secure and Safe AI'' Mobility Grant.

{
    \small
    \bibliographystyle{unsrtnat}
    \bibliography{main}
}

\newpage
\appendix
\input{appendix/main.tex}


\end{document}

%% file: preamble.tex
\usepackage[flushleft]{threeparttable}
\usepackage{capt-of}
\usepackage{wrapfig}

\usepackage{xspace}
\newcommand{\ours}{GRACE\xspace}
\definecolor{oursbg}{HTML}{E6F4E6}
\newcommand{\oursrow}{\rowcolor{oursbg}}

\newcommand{\ie}{i.e.\xspace}
\newcommand{\eg}{e.g.\@\xspace}

\usepackage{amssymb}
\usepackage{pifont}
\newcommand{\cmark}{\ding{51}}%
\newcommand{\xmark}{\ding{55}}%

\usepackage{amsmath}
\DeclareMathOperator*{\argmax}{\arg\!\max}

\newcommand{\splice}{SpLiCE\xspace}
\newcommand{\explainr}{\mbox{EXPLAIN-R}\xspace}

\usepackage{bm}
\newcommand{\vect}[1]{\bm{#1}}

%% file: sections/abstract.tex
\begin{abstract}
Vision-Language Models are designed to capture the compositional structure of the world through multi-modal alignment; however, it remains unclear whether their representations truly align with the way a human would naturally describe the same visual input.
Prior works attempt to decompose visual embeddings into their concept-level contributions, but their instance-level optimization is inherently local and does not capture the model’s global semantic structure, which is essential for \emph{compositional generalization} across different inputs, tasks, and domains.
We introduce GRACE, a method that learns this concept-based structure at the feature space level, producing semantic and general concept decompositions. 
To achieve this, we propose a novel training objective that preserves the geometric relations of different samples in the concept space, enabling a better understanding of how concepts compose and relate to each other.
We validate GRACE on several image and video benchmarks, showing that its concept decomposition is aligned with human perception and achieves faithful decompositions across different tasks and domains. With minimal overhead, \ours makes existing Vision-Language Models more interpretable, while preserving most of their original performance.
Project page with code at \href{https://sapeirone.github.io/grace/}{https://sapeirone.github.io/grace/}.
\end{abstract}

%% file: sections/introduction.tex
\section{Introduction}\label{sec:introduction}
Humans have the remarkable ability to perceive and understand reality by decomposing it into atomic semantic abstractions~\cite{fodor1988connectionism}.
In vision, few volumetric primitives are enough for our visual system to recognize familiar objects, even under occlusion or visual degradation~\cite{biederman1987recognition}. 
This mechanism, known as \emph{recognition-by-components}, forms the basis of human generalization. 
Natural language provides a powerful mechanism to express visual abstractions in a human-interpretable form and to enable their composition into rich descriptions of the world.
Thus, we expect \emph{interpretable} models to expose a reasoning structure that is well aligned with these abstractions, which is essential for \emph{compositional generalization}~\cite{wiedemer2023compositional}, \ie, the ability to recognize novel compositions of known~concepts.
\begin{figure*}[t]
    \centering
    \includegraphics[height=3.2cm, trim={1mm 1mm 1mm 1mm}]{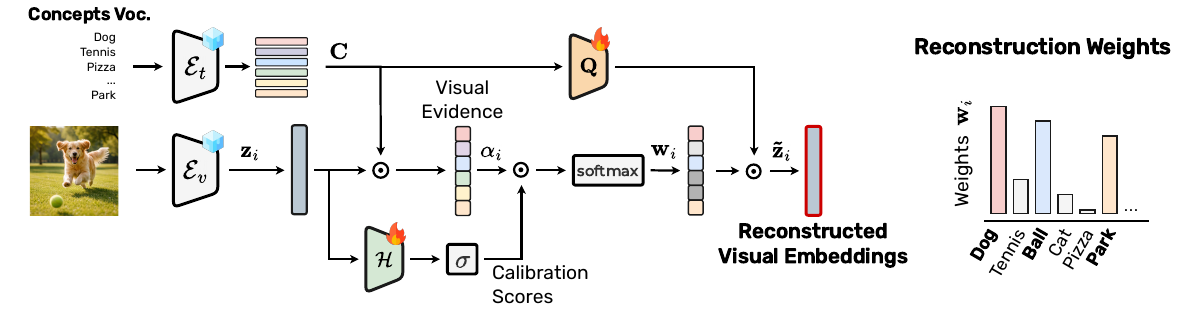}
    \caption{
        \textbf{The \ours framework.}
        Given a visual sample, either an image or video, and a vocabulary $\mathcal{C}$ of human-interpretable concepts, a geometry-based reconstruction pipeline decomposes the visual embedding into semantically relevant concept contributions. These reconstructed embeddings can replace the original ones, improving interpretability and enabling model debugging.
     }
    \vspace{-5mm}
    \label{fig:method_losses_architecture}
\end{figure*}
Vision–Language Models~\cite{radford2021learning} align their visual representations with this language-based compositional structure, achieving strong generalization~\cite{kempf2025when,mayilvahanan2024does}.
However, visual data are inherently complex and often encode multiple, tightly entangled, semantic concepts.
This challenge is particularly severe in videos, where objects, actions, and human motion co-occur and evolve over time~\cite{lee2025disentangled,fang2025humocon,knab2025concepts}.
As a result, it remains unclear whether their internal representations truly align with human-understandable concepts, which leads to a crucial question:
\begin{center}
    \emph{can we explicitly expose the interpretable concept-based structure of visual information in VLMs?}
\end{center}
In this regard, the linear hypothesis suggests that high-level concepts are already represented linearly in feature space~\cite{alain2016understanding,papadimitriou2025interpreting}, even across modalities~\cite{koishigarina2026clip}. 
Several works build on this intuition to decompose the visual embeddings of a VLM into text-based human-interpretable concepts.
\splice~\cite{bhalla2024interpreting} learns a \emph{per-sample} reconstruction of the embeddings from these concepts, while \explainr~\cite{hoang-xuan2025advancing} builds a concept-based surrogate representation that mimics the multi-modal alignment of the original model.
Yet, many challenges remain unsolved.
Instance-level decomposition may encourage shortcut learning and does not explicitly consider the different geometries of the visual and text modalities~\cite{goel2022cyclip}, while the surrogate representations in \explainr are obtained by collecting concept evidence from external captioning models~\cite{li2023blip}, rather than leveraging the intrinsic concept-based structure of the underlying VLM.
These models encode concept-level interpretability but lack a \emph{general} mechanism to expose it. 
Indeed, concept-based interpretability should capture not only the concept-level contributions within individual inputs, but also how concepts combine and relate across inputs, tasks, and domains, which is an essential property for generalization~\cite{kempf2025when}.

We address these challenges by introducing \ours, a novel concept-based framework that learns an interpretable decomposition of visual representations at the feature space level.
We leverage cross-modal matching over a large concept vocabulary to capture concept evidence directly from visual inputs, and propose a geometry-based reconstruction pipeline to ensure that the matched concepts are semantically relevant.
Rather than requiring exact reconstruction of the original visual embeddings, our method is designed to preserve the pairwise geometric relations between the samples in the concept space, which encourages the model to better understand how different concepts compose and relate to each other, thus better capturing the global semantic structure of the concept space.
Sparsity and consistency priors promote the selective use of relevant concepts and ensure consistent concept usage across semantically similar inputs.

We evaluate \ours across a range of benchmarks to show that its concept decompositions are well aligned with human perception. 
Its reconstructed embeddings preserve or even outperform the original embeddings, depending on the task, by exposing semantically relevant information and demonstrating strong \emph{zero-shot} generalization.
We summarize our contributions as follows: (i) we explore the connection between interpretability and generalization, addressing key limitations in existing approaches for concept-based decomposition; (ii) we propose \ours to expose the concept-based structure of VLMs, modeling how concepts compose within and across samples; (iii) we validate our method on several image and video benchmarks, showing its decomposition extracts sparse semantic concepts with strong generalization; (iv) \ours improves alignment with human perception by~\texttt{+}10.6\% on ConVisBench, while producing more faithful reconstructions (\texttt{+}7.50\% on EgoMCQ).
Our approach overcomes many limitations of similar previous approaches, and introduces an effective way to decompose visual embeddings into their concept-level contribution, improving interpretability while providing a practical tool to debug and better understand the model behavior.

%% file: sections/related_works.tex
\section{Background}

We consider different strategies to build models that expose human-understandable concepts.
In Concept Bottleneck Models, the raw input is explicitly mapped to high-level concepts, which are then used to solve the task, typically leveraging concept-level supervision.
Another line of research investigates how high-level concepts emerge as linearly separable directions in the representation space of VLMs, and how their contributions can be disentangled.
Our method brings together both views: it relies on vision-language models to match visual inputs with text-based human-understandable concepts, and learns a general concept-based decomposition of the embedding space.

\textbf{Concept Bottleneck Models.}
Concept Bottleneck Models (CBMs)~\cite{koh2020concept} decompose the internal representations of the model into a set of human-understandable concepts, and then learn a mapping function from the concept space to the output space of a specific target task, \eg, an image classification task. The bottleneck is trained with concept-level supervision obtained either from human annotations~\cite{koh2020concept} or by using LLMs~\cite{yang2023language,oikarinen2023labelfree} to generate plausible concept vocabularies. 
Recent works also apply CBMs to video tasks~\cite{knab2025concepts,lee2025disentangled}, learning to model the temporal evolution of concepts. While the bottleneck is commonly trained jointly with the downstream task, Post-hoc CBMs~\cite{yuksekgonul2023posthoc} extend concept-based interpretability to pretrained models by projecting their embeddings onto a set of interpretable activation vectors. 
CBMs learn task-specific mappings to the concept space, limiting the transferability of the learned representations across different tasks.

\textbf{Sparse Autoencoders.}
The \textit{linear representation hypothesis}~\cite{park2024the,trager2023linear,merullo2024language,muttenthaler2023improving} suggests that high-level concepts are represented linearly in feature space. 
Under this principle, Sparse Autoencoders (SAE)~\cite{cunningham2023sparse} learn a linear decomposition of a model's internal activations into sparse, monosemantic latent features, following an unsupervised activation reconstruction process.
SAEs have emerged as a relevant tool for concept-based interpretability in language~\cite{cunningham2023sparse,bussmann2025learning,gao2025scaling} and multimodal models~\cite{kim2025revelio,rao2024discover,thasarathan2025universal,fel2025archetypal,choi2026conceptscope,pach2025sparse}. 
The number of latent features is set to greatly exceed the size of the input activations to encourage disentanglement. 
Latents are then assigned a human-interpretable label by inspecting their activation patterns across different samples, possibly using an LLM to generate such explanations~\cite{paulo2025automatically}. 
However, this approach shifts interpretability to the explanation phase, which may generate oversimplified or broad labels~\cite{ma2026revising}.
Some hybrid approaches combine SAE-based decomposition and CBM interpretability in a single framework. 
DN-CBM~\cite{rao2024discover} first decomposes CLIP visual embeddings into sparse representations and then assigns each feature the closest label from a given vocabulary. 
ConceptScope~\cite{choi2026conceptscope} exploits SAE-based decomposition to distinguish between target and bias concepts in the input.
Recent works have questioned whether SAEs detect truly monosemantic features, such that each latent corresponds to a single, clearly defined concept~\cite{ICLR2026_3c1fe56b}. 
Others question whether SAE features generalize across data scarcity, class imbalance, label noise, and covariate shift, or merely match the performance of simple baselines~\cite{kantamneni2025are}. 
G-SAE~\cite{harle2026measuring} introduces a conditioning objective to encourage monosemanticity. 
Yet, even this approach requires ground-truth concept labels.

\textbf{Concept-based interpretability in VLMs.}
Recent works~\cite{bhalla2024interpreting,hoang-xuan2025advancing,yamaguchi2025zero} explore an interpretability-first approach by exploiting the linear representation hypothesis across modalities in CLIP-like models~\cite{koishigarina2026clip,papadimitriou2025interpreting}. These works first define a vocabulary of semantic concepts through language, enabling direct human-aligned interpretability, and then leverage a bottleneck-like approach to project the embeddings into the concept space and reconstruct them back. These works place strong emphasis on the faithfulness of the reconstructed representations, \ie, whether the reconstruction preserves the capabilities of the original embeddings on the downstream tasks. Even though they demonstrate the linearity of VLMs in concept space across modalities, they suffer from key limitations. The resulting mappings are either local, relying on instance-level decomposition~\cite{bhalla2024interpreting,yamaguchi2025zero}, which may result in shortcut learning, or leverage captioning models for concept detection~\cite{hoang-xuan2025advancing}.

GRACE learns a task-agnostic decomposition of visual embeddings that does not require concept-level supervision, leveraging the internal cross-modal structure of VLMs. Human-aligned interpretability of the decomposition is built into the model by design, as we force the model to reconstruct its features from a sparse set of predefined concepts, avoiding ambiguous post-hoc explanations. Our method integrates principles from SAE and CBM approaches to learn a general concept-based decomposition of the embedding space matching text-based and human-understandable concepts with visual inputs.

%% file: sections/method.tex
\section{Methodology}\label{sec:method}
We introduce \ours, a concept-based framework that decomposes the visual embedding space of any vision-language model into a combination of human-interpretable textual concepts, 
while maintaining multi-modal alignment and preserving the geometry of the samples in the concept space, \ie, the pairwise relations between the visual samples should remain the same when mapping through the concept space.
We address key limitations in prior work by extracting concept evidence directly from the visual inputs, thereby exposing the inherent concept-based structure of the input embeddings, and learning a general and task-agnostic decomposition that reasons about concept composition within and across different samples.
Our method learns a concept decomposition that reflects strong semantic relationships rather than task- or sample-specific cues, enabling effective transfer of the learned concept structure across tasks, datasets, and domains.
We illustrate \ours in Figure~\ref{fig:method_losses_architecture}.

\textbf{Problem definition.}
We denote by $\mathcal{E}_v$ and $\mathcal{E}_t$ the vision and text encoders of a pretrained vision-language model, mapping the two modalities in a shared~$\mathbb{R}^d$ space. 
We treat $\mathcal{E}_v$ as the \emph{target model}, \ie, the model we aim to make interpretable, and indicate as $\mathbf{z}_i = \mathcal{E}_v(x_i)$ the visual embedding of any sample $x_i$.
$\mathcal{C}$ is a large vocabulary of $K$ textual concepts, representing individual semantic elements, \eg, simple actions or objects, that may appear in the visual input.
We look for a mapping~$h\colon \mathbb{R}^d \to \mathbb{R}^K$, such that~$h \circ \mathcal{E}_v$ yields a decomposition encoding the relative contribution of the concepts in any given visual input.
To ensure that the matched concepts represent relevant semantic information for the input, we introduce a mapping $r\colon \mathbb{R}^K \to \mathbb{R}^d$ that maps back to the original embeddings and minimize a reconstruction objective $\ell(\mathbf{\tilde{z}}_i, \mathbf{z}_i)$ between the original visual embeddings $\mathbf{z}_i$ and their reconstruction~$\mathbf{\tilde{z}}_i = (r \circ h)(\mathbf{z}_i)$.
The choice of $\ell$ is critical, as it determines \emph{how} and \emph{how much} of the original embeddings is preserved through the concept-based decomposition. 

\textbf{Intuition.}
For any given visual sample, we detect initial concept evidence by matching its visual embedding against the text embeddings of a predefined concept vocabulary through cross-modal cosine similarity. 
We then look for a mapping back to the original visual embeddings by learning the optimal reconstruction weights from the visual evidence.
Prior works enforce strict $\ell_2$ reconstruction~\cite{bhalla2024interpreting}, which is fundamentally limited by the modality gap and the different geometries of the two modalities~\cite{liang2022mind,goel2022cyclip}. 
Instead, we opt for a relaxed reconstruction objective that preserves the geometry of the reconstructed visual embeddings in concept space, \ie, the pairwise similarity of features before and after reconstruction.
This approach goes beyond instance-level reconstruction and encourages the model to better understand how concepts are composed within and across samples, improving compositional reasoning.
We push the reconstruction weights to capture rich and meaningful semantic information from the input, while promoting (i) \emph{sparsity}, \ie, samples should be reconstructed by few concepts, and (ii) \emph{consistency}, \ie, reconstruction weights should be consistent across semantic~categories.

\begin{figure}[t]
    \centering
    \includegraphics[width=0.9\linewidth]{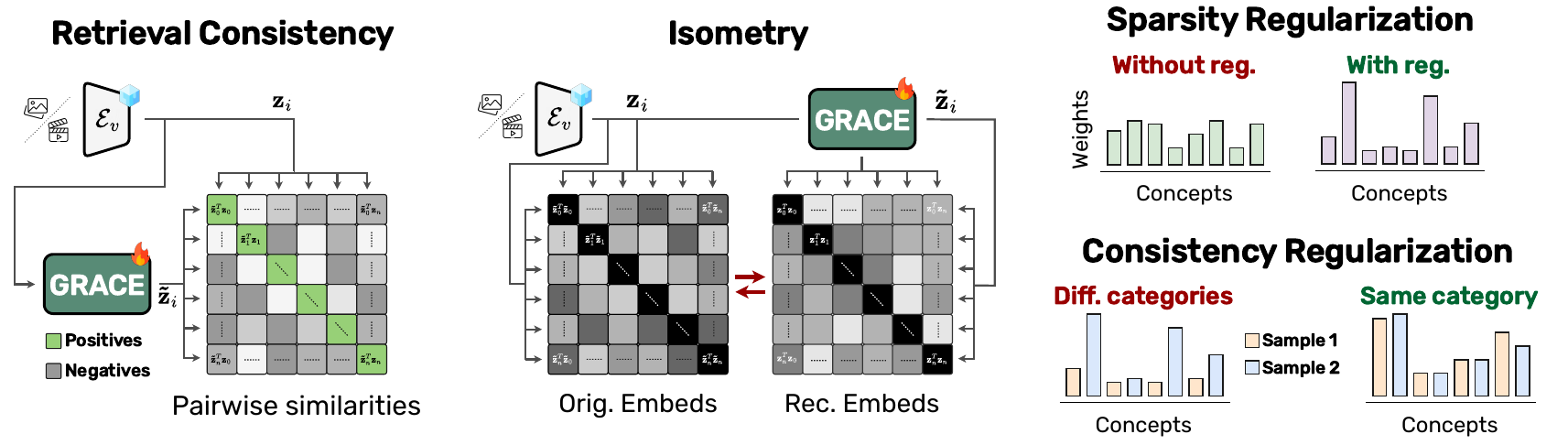}
    \vspace{-1mm}
    \caption{\textbf{Geometry-based reconstruction and regularization objectives in \ours.} \emph{Retrieval consistency} ensures that the reconstructed visual embeddings are close to their original counterparts, \emph{Isometry} preserves the same pairwise similarities between the samples in the batch before and after the reconstruction, while \emph{Sparsity} and \emph{Consistency} regularization favor sparse reconstruction weights and consistency across different sample categories. 
    }
    \label{fig:method_losses_intuition}
    \vspace{-1em}
\end{figure}

\subsection{Geometry-based reconstruction}\label{sec:method_concepts_based_reconstruction}

\noindent\textbf{Collecting visual concept evidence.}
We exploit the cross-modal \emph{bag-of-words} nature of vision-language models~\cite{yuksekgonul2023when,bhalla2024interpreting,koishigarina2026clip} to detect individual concepts from the vocabulary in a given visual sample by measuring the cosine similarity between the visual embeddings $\mathbf{z}_i$ and the text embedding of the concept $\mathbf{c}_j$. We refer to these similarities as \emph{visual evidence} $\alpha_{i}$, computed as $\alpha_{i,j} = \cos(\mathbf{z}_i, \mathbf{c}_{j})$.
Concept vocabularies can be obtained from multiple sources, including human supervision~\cite{koh2020concept}, LLMs~\cite{yang2023language,tan2024explain}, or by mining frequent terms from captions. 
In this work, we adopt the mining strategy for simplicity and scalability, even though \ours is robust to the selection of the concepts vocabulary.
However, synonyms in the vocabulary can skew the visual evidence toward uniformly positive values, resulting in many detected concepts. 
To mitigate this phenomenon, we introduce a gating mechanism similar to GLU~\cite{dauphin2017language} that calibrates the relevance of each concept for the given input. 
We define the \emph{calibrated evidence} $\tilde{\alpha}_{i}$~as:
\begin{equation}\label{eq:calibrated_acts}
    \tilde{\alpha}_{i} = \alpha_{i} \odot \mathcal{H}(\mathbf{z}_i),
\end{equation}
where $\mathcal{H}$ is an MLP with a sigmoid activation function on the output.
Desirably, the calibrated evidence should preserve only a few concepts per sample, with a long tail of concepts with near-zero weight, directly improving interpretability and the semantic relevance of the matched concepts.

\noindent\textbf{Visual embeddings reconstruction.}
We ensure that the calibrated evidence reflects meaningful concepts by using them to reconstruct the original embeddings. 
Prior works directly minimize the $\ell_2$ distance between the reconstructed and original embeddings subject to a sparsity penalty on the reconstruction coefficients~\cite{bhalla2024interpreting}, \eg, $\min_{\mathbf{w}_i} ||\mathbf{z}_i - \mathbf{w}_i^T \mathbf{C}||_2^2 + ||\mathbf{w}_i||_1$, or learn a \emph{surrogate} concept-based representation $\mathbf{\tilde{z}}_i$ that preserves the multi-modal alignment of the original model~\cite{hoang-xuan2025advancing}, \eg, $\cos(\mathbf{\tilde{z}}_i, \mathbf{t}_j) \approx \cos(\mathbf{z}_i, \mathbf{t}_j)$ for any possible combination $(i,j)$ of visual and text samples in the batch. 
However, these approaches overlook the fact that the visual and text representations are not interchangeable and have different geometries~\cite{liang2022mind,goel2022cyclip}, which prevents a full reconstruction of the visual embeddings as a linear combination of text-encoded concepts. 
Instead, we relax such a strict reconstruction process using a simpler geometry reconstruction objective, which preserves the pairwise similarities between the visual samples in the batch when mapped to the concept space beyond instance-level reconstruction.
Formally, we compute our \emph{reconstructed visual embeddings}~as:
\begin{equation}\label{eq:reconstruction}
    \mathbf{\tilde{z}}_i = \sigma (\tilde{\alpha}_i^T / \tau)^T \mathbf{C}\mathbf{Q} = \mathbf{w}_i^T \mathbf{C}\mathbf{Q},
\end{equation}
where $\sigma$ is the softmax function over the evidence, $\tau$ is a temperature parameter and $\mathbf{Q}\in \mathbb{R}^{d\times d}$ is a learnable orthogonal transformation, \ie, $\mathbf{Q}\mathbf{Q}^T=\mathbb{I}$, of the concept embeddings.
The orthogonal transformation maps the concepts in a subspace compatible with the visual embeddings without further entangling them, \ie, the dot product before and after the projection is preserved $(\mathbf{Q}^T \mathbf{c}_i)^T(\mathbf{Q}^T \mathbf{c}_j) = \mathbf{c}_i^T \mathbf{c}_j$. 
We discuss the importance of this property for interpretability in Sec.~\ref{sec:appendix_projection}.

\noindent\textbf{Geometry reconstruction.}
We define a dual \emph{geometry reconstruction} objective that ensures: (i) \emph{retrieval consistency}, such that the reconstructed embeddings are ranked correctly against the original embeddings, and (ii) \emph{isometry}, enforcing that the pairwise cosine similarities of the reconstructed embeddings match the original representations. We illustrate these objectives in Figure~\ref{fig:method_losses_intuition}.
For retrieval consistency, we use a contrastive loss in which reconstructed embeddings are pushed closer to the original counterparts, maximizing their mutual information, and further apart from other samples in the batch. 
For isometry, we minimize the distance between the cross-similarity matrices of the original and reconstructed embeddings: 
\begin{equation}\label{eq:rec}
    \scalebox{0.95}{$\mathcal{L}_{rec} = \underbrace{\sum_{\mathbf{z}_i \in \mathcal{B}} \left[ -\log \frac{e ^ {\mathbf{\tilde{z}}_i^T \mathbf{z}_i / \tau}}{\sum_{k} e^{\mathbf{\tilde{z}}_i^T \mathbf{z}_k / \tau}} \right]}_{\text{Retrieval consistency}} + \underbrace{|| \mathbf{\tilde{Z}}\mathbf{\tilde{Z}}^T - \mathbf{Z}\mathbf{Z}^T||_F^2}_{\text{Isometry}},$}
\end{equation}
where $\mathbf{Z},\mathbf{\tilde{Z}} \in \mathbb{R}^{|\mathcal{B}|\times d}$ are the $\ell_2$-normalized original and reconstructed visual embeddings in the batch. 
The \emph{retrieval consistency} term enforces that, for any reconstructed embedding~$\mathbf{\tilde{z}}_i$, its closest match among the original embeddings is~$\mathbf{z}_i$, \ie, $\argmax_j \cos(\mathbf{\tilde{z}}_i, \mathbf{z}_j) = i$.
The \emph{isometry term} ensures that the cosine similarity between any two reconstructed samples is close to that of their original embeddings, \ie, $\text{cos}(\mathbf{\tilde{z}}_i, \mathbf{\tilde{z}}_j) \approx \text{cos}(\mathbf{z}_i, \mathbf{z}_j)$.
Together, these two objectives preserve the unimodal pairwise relations of the original and reconstructed embeddings, capturing the global semantic structure of the model, without imposing strict instance-level reconstruction.

\subsection{Sparsity and consistency properties}\label{sec:method_regularizations}
Our reconstruction objective does not impose any constraint on the reconstruction weights~$\mathbf{w}_i$. 
Yet, we argue that a \emph{good} concept-based decomposition should satisfy two key properties: \emph{sparsity} and \emph{consistency}.
Visual samples should be compressed into a few significant concepts, thereby improving interpretability, while promoting consistency in the reconstruction weights across different categories, \eg, food-related concepts are more likely to appear in images or videos set in a \emph{kitchen}, but less likely in \emph{sport} videos. We discuss in Sec.~\ref{sec:exp_regularization_analysis} that this property emerges naturally from the~data.

\noindent\textbf{Sparsity.} 
We parametrize the reconstruction weights to follow a categorical distribution~$\text{Cat}(\mathbf{w}_{i})$.
To explicitly control the sparsity of the reconstruction process, we assume $\mathbf{w}_{i}$ from Eq.~\ref{eq:reconstruction} to be drawn from a symmetric Dirichlet distribution $\mathbf{w}_i \sim \text{Dir}(\omega)$, whose density is proportional to $\prod_j \mathbf{w}_{i,j}^{\omega - 1}$ and $\omega$ is a concentration parameter in $(0,1]$. 
For lower values of $\omega$, sparse reconstruction weights are more likely, while higher values favor more uniform weights.
The concentration parameter $\omega$ acts as a tunable hyperparameter that controls the sparsity of the reconstruction. 
The model is trained to minimize the negative log-likelihood of the reconstruction weights under the Dirichlet prior:
\begin{equation}
    \mathcal{L}_{reg} = - \sum_{i=1}^{|\mathcal{B}|} \left[ (\omega - 1) \sum_{k=1}^{K} \log{(\mathbf{w}_{i,k} + \epsilon)} \right],
\end{equation}
where $\epsilon$ prevents instabilities as $\mathbf{w}_{i,k}\to0$.

\noindent\textbf{Consistency.} 
We encourage the model to predict similar concepts for semantically similar samples, \eg, videos from the same location.
This approach builds on prior work in disentangled representations which seek to isolate shared and unique latent factors across samples and modalities~\cite{von2021self,fumero2023leveraging,liang2023factorized}.
We encourage the distribution of the reconstruction weights $(\mathbf{w}_i, \mathbf{w}_j)$ to be similar for samples $(i,j)$ with overlapping categories, \eg, videos captured in the same location, and dissimilar otherwise:
\begin{equation}\label{eq:jsd}
    \mathcal{L}_{con} = \sum_{i=1}^{|\mathcal{B}|}\sum_{j=1}^{|\mathcal{B}|} \delta_{ij} \text{D}(\mathbf{w}_{i}||\mathbf{w}_{j}),
\end{equation}
where $\delta_{ij}$ is $+1$ if samples $i$ and $j$ are from the same category and $-1$ otherwise. $\text{D}$ is the Jensen-Shannon Divergence (JSD)~\cite{lin1991divergence} between the reconstruction weights of the two samples and is defined as $\text{D}(p||q) = \frac{1}{2} \left[ \text{KL}(p||m) + \text{KL}(q||m)\right]$, with $m = \frac{1}{2}(p+q)$. 

\subsection{Training and inference}
We train \ours with a combination of these objectives, $\mathcal{L} = \mathcal{L}_{rec} + \lambda_r \mathcal{L}_{reg} + \lambda_c \mathcal{L}_{con}$, where $\lambda_r$ and $\lambda_c$ control the weight of the regularization terms, and are chosen to minimize the reconstruction objective on the training data.
We first train \ours on large-scale image and video datasets to learn a general task-agnostic concept decomposition, which can then be directly applied to unseen tasks and domains.
At inference time, the reconstructed embeddings~$\mathbf{\tilde{z}}_{i}$ can be used in place of the original embeddings, while the reconstruction weights~$\mathbf{w}_{i}$ provide an interpretable bottleneck encoding the relative contribution of each concept.

%% file: sections/experiments.tex
\section{Experiments}\label{sec:experiments}

\input{sections/tables/convisbench.tex}
\input{sections/tables/video_retrieval}

We consider a diverse set of image and video benchmarks to evaluate \ours.
With \emph{Concept-based Similarity Estimation}, we measure how the reconstruction weights align with human perception, capturing semantic concepts from the input.
With \emph{classification} and \emph{retrieval} tasks, we demonstrate that the interpretable decomposition from \ours generalizes well across different tasks and domains.
Together, these benchmarks show the connection between interpretability and generalization in VLMs.

\subsection{Video Benchmarks}
We train \ours on EgoClip~\cite{lin2022egocentric}, a collection of 3.8M video-text pairs derived from Ego4D~\cite{grauman2022ego4d}.
Then, we evaluate our method on text-to-video retrieval on \textbf{EgoMCQ}~\cite{lin2022egocentric}, action recognition on \textbf{Charades-Ego}~\cite{sigurdsson2018charades} and video similarity estimation on \textbf{ConVisBench}~\cite{liberatori2025convisbench}, a task where the goal is to determine the semantic similarity of two given videos.
Together, we use these benchmarks to demonstrate that our reconstructed embeddings preserve vision-language alignment and generalize across different tasks and domains, while capturing rich semantic information.
We use the Perception-Encoder~\cite{bolya2025perception} backbone and mine the concept vocabulary from the EgoClip captions. For the consistency regularization term, we use the scenario labels from Ego4D.

\textbf{Concept-based Similarity Estimation.} 
In ConVisBench, the goal is to predict whether two videos are similar with respect to a predefined semantic axis, \eg, the \emph{location} in which the video was recorded or the \emph{main subject}.
Performance is measured as Spearman’s and Kendall’s correlation between the predicted similarity score and the human similarity annotation for each pair.
Notably, all the previous methods evaluated on this benchmark are not explicitly concept-based.
\emph{Text-to-Text} methods measure the similarity of the video captions with SBERT~\cite{reimers2019sentence}, while \emph{Video-to-Video} methods use the cosine similarity between global video embeddings.
\emph{Cross-modal} methods evaluate similarity using CLIPScore~\cite{hessel2021clipscore}.
For \ours, we compute similarity using either CLIPScore on the reconstructed embeddings or the distance between the two samples' reconstruction weights.
Notably, the latter approach achieves the highest correlation with human judgment, demonstrating that the reconstruction weights capture semantically meaningful concepts in the videos, as shown in Table~\ref{tab:ConVisBench}.
\ours also outperforms the non-interpretable baseline, showing that the concept bottleneck isolates semantically relevant information while filtering out irrelevant noise.

\textbf{Video Retrieval and Action Recognition.} 
\textbf{EgoMCQ} consists of 39k multiple-choice \emph{text-to-video} questions where, given a textual query, the model has to select the correct video clip among five candidates.
We report Inter-Video accuracy, where clips are from different videos, and Intra-Video accuracy, where they are from the same video.
We use cosine similarity between the reconstructed visual embeddings and the text embedding of the query to find the best match.
\textbf{Charades-Ego} is a dataset of 7.8K videos with annotated actions from a taxonomy of 157 classes. Performance is evaluated using mAP.
We define a zero-shot classifier using the text embeddings of the action labels.
For both benchmarks, we also evaluate faithfulness, \ie, measuring how well the reconstructed embeddings preserve the predictions of the original embeddings, as defined by~\cite{hoang-xuan2025advancing}.
As shown in Table~\ref{tab:video_retrieval}, \ours consistently improves on faithfulness, despite not relying on instance-level optimization or caption-based evidence as in \splice and \explainr, respectively.
For a fair comparison, we also integrate the multi-modal distillation objective from~\cite{hoang-xuan2025advancing}.
This addition provides marginal gains, showing that our uni-modal reconstruction is sufficient to preserve multi-modal~alignment.

\subsubsection{Analyses of the concept decomposition}\label{sec:exp_regularization_analysis}

\noindent\textbf{Sparsity of the decomposition.} 
We define the \emph{cumulative reconstruction mass at $k$ (CRM@k)} as the sum of the top-$k$ largest reconstruction weights, and report its trend in Figure~\ref{fig:crm} as the sparsity parameter $\omega$ varies.
Low values of $\omega$ reduce the number of non-zero reconstruction weights.
\ours achieves near-perfect reconstruction with few concepts, thanks to its general reconstruction objective.

\noindent\textbf{Disentanglement of the reconstruction weights.} 
We adapt the DCI metrics~\cite{eastwood2018a} to our setting, by training a set of sparse logistic regression classifiers to predict the video scenario, \ie, a coarse label describing the main activity in the video, from just the reconstruction weights, and evaluate their behavior in terms of (i) \emph{sparsity}, \ie, how many concepts are required for the prediction (L\textsubscript{0} Norm); (ii) \emph{disentanglement}, measuring the separation of concepts across scenarios; (iii) \emph{completeness}, reflecting whether each scenario is explained by few concepts; and (iv) \emph{informativeness}, corresponding to prediction accuracy (mAP).
As shown in Table~\ref{tab:disentanglement}, \splice achieves good prediction accuracy but poor disentanglement and sparsity, indicating substantial overlap in the concepts matched across different scenarios.
In contrast, \ours is slightly less accurate but exhibits stronger disentanglement, indicating that scenario-level disentanglement already emerges from the data. More details in Sec.~\ref{sec:appendix_disentanglement}.

\noindent\textbf{Concept redundancy and sufficiency.} 
We consider the Singular Value Decomposition of the concept matrix to estimate redundancy in the vocabulary. 
Let $s_i$ denote the singular values of $\mathbf{C}$ in decreasing order. We define $k$ as the smallest integer such that $\sum_{i=1}^k s_i^2 / \sum_{i=1}^K s_i^2 \ge \tau$, where $\tau$ is a threshold hyperparameter that controls the amount of retained concept variance, and thus the completeness of the selected concepts.
Then, we perform a QR factorization with column pivoting of the concept matrix, iteratively selecting the concept with the largest residual contribution. The retained concept embeddings form a set of \emph{sufficient} concepts, based on the choice of $\tau$.
The concept matrix can be heavily compressed into few concepts, without significant performance degradation (Table~\ref{tab:vocabulary}-left).

\noindent\textbf{Construction of the concept vocabulary.}
We show in Table~\ref{tab:vocabulary}-right the impact on performance of different concept vocabularies, including one obtained by prompting GPT-5.4 for an exhaustive list of plausible text-based concepts that could appear in videos of human activities. This approach assumes little knowledge about the videos and minimal human effort, yet it achieves good performance. 

\input{sections/tables/crm_dci.tex}
\input{sections/figures/vocabulary.tex}

\input{sections/tables/image_retrieval_faithfullness.tex}

\input{sections/tables/image_cls_faithfullness.tex}
\input{sections/figures/shift_bias.tex}

\input{sections/figures/qualitatives}
\subsection{Image benchmarks}

We evaluate our method on several tasks, including \textbf{COCO}~\cite{lin2014microsoft} and \textbf{Flickr30k}~\cite{plummer2015flickr30k}, \textbf{ImageNet}~\cite{deng2009imagenet}, \textbf{Flower102}~\cite{flowers102} and \textbf{Food101}~\cite{food101}, measuring task performance and faithfulness.
We train \ours on ImageNet and evaluate it on these tasks without fine-tuning, using the CLIP ViT-B/32 backbone.
We adapt \explainr to the same training and validation protocol, and report its original performance when dataset-specific training is available.
We use a vocabulary mined from ImageNet captions, and compare different vocabularies in Sec.~\ref{sec:appendix_vocabulary}.
As for video tasks, we also consider a variant of \ours that leverages captions during training to improve multi-modal~alignment.

\noindent\textbf{Image-Text Retrieval.} 
\ours produces significantly higher-quality reconstructed embeddings, while being competitive with previous methods on the task metrics (Table~\ref{tab:image_retrieval}).
Our reconstructed embeddings are much closer to the original embeddings in terms of multi-modal alignment (\emph{higher faithfulness}), even though the model is slightly less confident in the relative ranking of the image-text pairs, especially at lower recall thresholds.
We attribute this gap to the use of caption-derived concept evidence in \explainr, which simplifies vision–language alignment.

\noindent\textbf{Image Classification.} 
Our approach consistently outperforms \splice and \explainr in both accuracy and faithfulness, despite not having access to caption-based evidence and not using instance-level optimization (Table~\ref{tab:image_classification}).
\explainr suffers substantial performance degradation in the zero-shot setting, whereas the geometry-preserving reconstruction process of \ours yields a more transferable concept decomposition that generalizes across domains.
This method leverages an external captioning model to extract concept evidence, which limits its applicability.
In contrast, our method avoids this assumption by learning the most relevant concepts directly from visual features.

\subsubsection{Robustness to severe concept shift and bias analysis}
We define five different folds over the ImageNet classes, such that the overlap between the train and test classes is minimized.
Completely disjoint concept sets are not possible, as some common concepts are ubiquitous and occur frequently in many classes.
We train \ours on the train set of each fold and evaluate zero-shot on the corresponding test set, comparing with the model trained on all the classes. 
Even though the reconstructed embeddings are tested on zero-shot classes, \ours maintains performance very close to the model trained on all classes, as shown in Figure~\ref{fig:shift}.
These experiments demonstrate the robustness of \ours to severe shifts in the concept distribution across training and evaluation splits.
A \ours model trained on a set of classes with a given class-concept distribution can still produce reasonable decompositions when tested on classes with a very significant shift in their class-concept distributions.

\noindent\textbf{Bias analysis.}
We analyze the average top-5 concept activations for a subset of the ImageNet classes, showing some interesting patterns in Table~\ref{tab:bias_analysis}. Locations, e.g., ``\textit{chile}'' and ``\textit{indonesia}'', often appear in the volcano class, showing a strong entity-location correlation. 
Similarly, ``\textit{tokyo}'' is the strongest activated concept in the rickshaw class. ``\textit{casio g}'' is often activated for devices featuring digital displays, even though these objects are not necessarily watches.

\subsection{Qualitative results}
We compare the concepts activated by SpLiCE and \ours on ConVisBench videos.
Both methods recover meaningful concepts for each video, but \splice tends to concentrate most of the weights in the very first few concepts, which is a direct consequence of its instance-level optimization.
Figure~\ref{fig:qualitatives_interventions} shows that this behavior makes \splice sensitive to concept interventions, \ie, when some matched concepts are artificially removed to simulate prediction errors.
In this setting, we consider a pair of videos from ConVisBench and compute their reconstruction weights.
We zero out the weight of a predicted concept, rescale the remaining ones, and measure similarity before and after the intervention.
\ours is only slightly affected by the intervention, as the similarity score between the two videos remains almost the same, while \splice exhibits a more significant drop, even though they remain similar in terms of location, main subject, and objects.
More details in Sec.~\ref{sec:appendix_qualititatives}.

%% file: sections/tables/convisbench.tex

\begin{table*}[t]
    \scriptsize
    \centering
    \setlength{\tabcolsep}{3.5pt}
    \caption{
        \textbf{Concept-based Similarity Estimation on ConVisBench.}
        Given a video pair, the model has to predict their semantic similarity across different axes, \eg, \emph{Main Objects} or \emph{Location}.
        Performance is measured by Spearman’s $\rho$ and Kendall’s $\tau$ correlations against human judgment.
    }\label{tab:ConVisBench}
    \vspace{-2mm}
    \begin{tabularx}{\textwidth}{Xlccccc|c}
        \toprule
        \multirow{2}{*}{\textbf{Model}} & \multirow{2}{*}{\textbf{Method}} 
            & \textbf{Main Action}
            & \textbf{Main Subjects}
            & \textbf{Main Objects}
            & \textbf{Location}
            & \multicolumn{1}{c}{\textbf{Order of Actions}} & \textbf{Average} \\
        &   & \multicolumn{1}{c}{$\rho$ / $\tau$} 
            & \multicolumn{1}{c}{$\rho$ / $\tau$} 
            & \multicolumn{1}{c}{$\rho$ / $\tau$} 
            & \multicolumn{1}{c}{$\rho$ / $\tau$} 
            & \multicolumn{1}{c}{$\rho$ / $\tau$}
            & \multicolumn{1}{c}{$\rho$ / $\tau$} \\
        \midrule
        \multicolumn{7}{l}{\textcolor{gray}{\emph{\textbf{Text-to-Text}}}}\\
        mPLUG-Owl3~\cite{ye2025mplugowl} & SBERT 
            & 52.1 / 36.2
            & 45.5 / 31.1
            & 55.1 / 38.9
            & 28.4 / 19.6
            & 49.9 / 34.9
            & 46.2 / 32.1
        \\
        InternVL2.5~\cite{chen2024expanding} & SBERT 
            & 39.4 / 27.4
            & 31.2 / 21.4
            & 45.0 / 31.1
            & 29.9 / 20.5
            & 42.0 / 29.5
            & 37.5 / 26.0
        \\
        Qwen2.5-VL (7B)~\cite{bai2025qwen2} & SBERT 
            & 42.3 / 29.5
            & 45.4 / 31.7
            & 45.4 / 31.8
            & 38.5 / 26.7
            & 40.9 / 28.4
            & 42.5 / 29.6
        \\
        Qwen3.5 (9B)~\cite{qwen35}	& SBERT	
            & 36.5 / 25.3
            & 37.1 / 25.8
            & 42.9 / 30.0
            & 34.8 / 23.9
            & 37.0 / 25.6
            & 37.6 / 26.1
        \\
        \midrule
        \multicolumn{7}{l}{\textcolor{gray}{\emph{\textbf{Video-to-video}}}} \\
        VideoMAE~\cite{tong2022videomae} & Cosine Sim. 
            & 13.0 / 8.7
            & 23.1 / 15.6
            & 13.2 / 8.8
            & 37.8 / 26.4
            & 15.1 / 10.2
            & 20.4 / 14.0
        \\
        DINOv2~\cite{oquab2024dinov} & Cosine Sim. 
            & 33.3 / 22.6
            & 40.9 / 28.6
            & 37.4 / 25.8
            & 57.4 / 41.7
            & 34.6 / 24.0
            & 40.7 / 28.5
        \\
        \midrule
        \multicolumn{7}{l}{\textcolor{gray}{\emph{\textbf{Cross-modal}}}} \\
        CLIP~\cite{radford2021learning} & CLIPScore 
            & 35.7 / 24.6
            & 39.0 / 27.0
            & 39.0 / 27.1
            & 36.0 / 24.9
            & 30.9 / 21.1
            & 36.1 / 24.9
        \\
        InternVideo~\cite{wang2022internvideo} & CLIPScore 
            & 43.1 / 30.1
            & 54.8 / 38.7
            & 54.5 / 38.5
            & 40.9 / 28.4
            & 40.2 / 28.1
            & 46.7 / 32.8
        \\
        LLaVA-OV~\cite{li2025llavaonevision} & VQAScore 
            & 51.1 / 36.1
            & 55.8 / 39.6 
            & 58.3 / 41.4
            & 46.5 / 32.3
            & 48.1 / 33.9
            & 52.0 / 36.6 
        \\
        Perception-Encoder~\cite{bolya2025perception} & CLIPScore
            & 51.1 / 36.0
            & 57.9 / 41.3
            & 57.5 / 41.2
            & 44.4 / 31.3
            & 47.6 / 33.5
            & 51.7 / 36.7
        \\
        \midrule
        \midrule
        \multicolumn{7}{l}{\textcolor{gray}{\emph{\textbf{Concept-based methods (using Perception-Encoder backbone)}}}}\\
        SpLiCE~\cite{bhalla2024interpreting} & CLIPScore 
            & 50.6 / \underline{35.8}
            & 41.0 / 28.0
            & 53.0 / 37.4
            & 31.9 / 21.7
            & 47.0 / 32.8
            & 44.7 / 31.1
        \\
        %
        %
        EXPLAIN-R~\cite{hoang-xuan2025advancing}  & CLIPScore 
            & 41.1 / 28.8
            & 29.9 / 20.2
            & 44.9 / 31.3
            & 29.9 / 20.5
            & 40.0 / 27.7
            & 37.2 / 25.7
        \\
        %
        \oursrow \textbf{\ours} & CLIPScore 
            & 43.7 / 30.6
            & 47.0 / \underline{32.9}
            & 44.3 / 31.2
            & 32.0 / 22.1
            & 40.3 / 27.9
            & 41.5 / 28.9
        \\
        \midrule
        SpLiCE~\cite{bhalla2024interpreting} & Weights (L1) 
            & \underline{51.3} / 35.4
            & \underline{47.7} / 32.8
            & \textbf{62.1} / \textbf{44.1}
            & \underline{43.7} / \underline{29.9}
            & \underline{49.5} / \underline{34.5}
            & \underline{50.9} / \underline{35.4}
        \\
        \oursrow \textbf{\ours} & Weights (JSD)  
            & \textbf{54.0} / \textbf{38.2}
            & \textbf{60.9} / \textbf{43.5}
            & \underline{58.6} / \underline{42.2} 
            & \textbf{56.3} / \textbf{40.1}
            & \textbf{51.6} / \textbf{36.6}
            & \textbf{56.3} / \textbf{40.1}        
        \\
        \bottomrule
    \end{tabularx}
    \vspace{-4mm}
\end{table*}

%% file: sections/tables/video_retrieval.tex
\begin{table}[t]
    \scriptsize
    \centering
    \caption{
        \textbf{Accuracy and Faithfulness on video retrieval (EgoMCQ) and action recognition (Charades) tasks.}
        \ours and \explainr are trained on EgoClip and evaluated in \emph{zero-shot}. 
    }\label{tab:video_retrieval}
    \begin{tabularx}{0.75\linewidth}{Xccccccc}
        \toprule
        \multirow{2}{*}{\textbf{Method}}                    & \multirow{2}{*}{\textbf{Zero-Shot}} & \multicolumn{3}{c}{\textbf{EgoMCQ}} & \multicolumn{2}{c}{\textbf{Charades}}                                                             \\
                                                            &                                     & Intra Acc.                          & Inter Acc.                            & Faith.            & Acc.              & Faith.            \\
        \midrule
        Perception Encoder~\cite{bolya2025perception}       & \cmark                              & 37.64                               & 84.88                                 & N/A               & 31.30             & N/A               \\
        \midrule
        \splice~\cite{bhalla2024interpreting}               & \cmark                              & 34.60                               & 79.04                                 & 70.36             & \textbf{26.94}    & 29.70             \\
        \explainr~\cite{hoang-xuan2025advancing}            & \cmark                              & 30.10                               & 76.17                                 & 63.70             & 21.73             & 25.92             \\
        \oursrow \textbf{\ours}                             & \cmark                              & \underline{35.43}                   & \underline{81.35}                     & \underline{77.86} & \underline{25.99} & \underline{35.15} \\
        \oursrow \textbf{\ours} (\texttt{+}\emph{distill.}) & \cmark                              & \textbf{35.50}                      & \textbf{82.94}                        & \textbf{80.48}    & 25.97             & \textbf{39.53}    \\
        \bottomrule
    \end{tabularx}
    \vspace{-4mm}
\end{table}

%% file: sections/tables/crm_dci.tex
\begin{table*}[t]
    \centering
    \begin{tabular}{@{}p{0.49\textwidth}@{\hspace{0.02\textwidth}}p{0.49\textwidth}@{}}
        \begin{minipage}[t]{\linewidth}
            \vspace{0pt}
            \centering
            \includegraphics[width=0.95\linewidth, trim={0 0.3cm 0 0.5cm}]{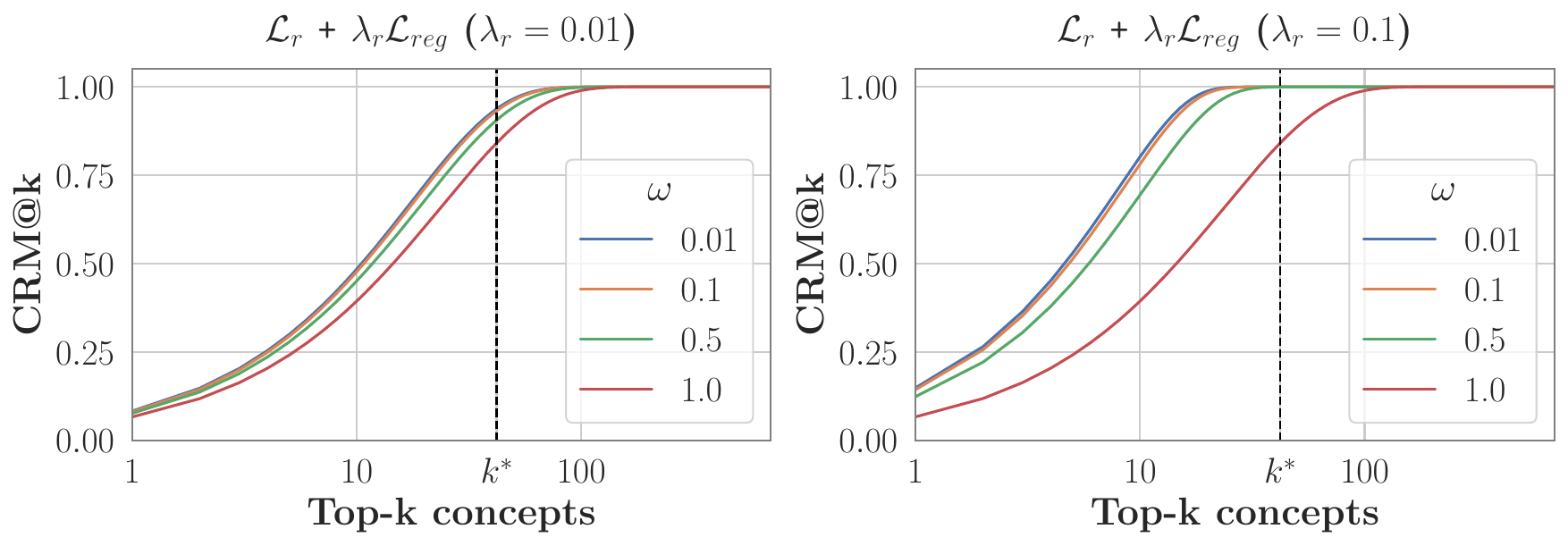}
        \end{minipage}
         &
        \begin{minipage}[t]{\linewidth}
            \vspace{0pt}
            \centering
            \scriptsize
            \setlength{\tabcolsep}{1pt}
            \vspace{2mm}
            \begin{tabular}{@{}lcccc@{}}
                \toprule
                \textbf{Method}                                    & \textbf{mAP} ($\uparrow$) & \textbf{Disent.} ($\uparrow$) & \textbf{Complet.} ($\uparrow$) & \textbf{L}\textsubscript{\textbf{0}} \textbf{Norm} ($\downarrow$) \\
                \midrule
                SpLiCE~\cite{bhalla2024interpreting}                                             & 86.1                      & 8.9                           & 18.3                           & 633.2                                                             \\
                \oursrow \textbf{\ours} ($\mathcal{L}_{rec}$)                                & 72.5                      & 53.4                          & 93.8                           & 28.5                                                              \\
                \oursrow \textbf{\ours} (\texttt{+}$\mathcal{L}_{con}$) & 73.3                      & 53.9                          & 93.8                           & 27.8                                                              \\
                \bottomrule
            \end{tabular}
        \end{minipage} \\[-1mm]
        \begin{minipage}[t]{\linewidth}
            
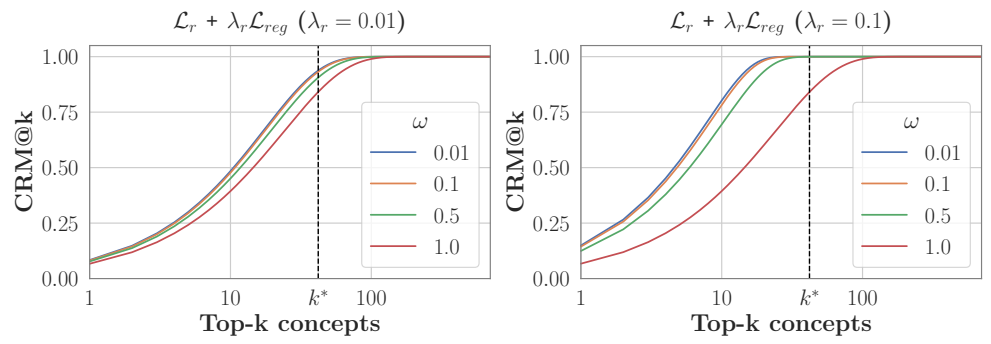
\captionof{figure}{
                \textbf{Cumulative Reconstruction Mass (CRM@k) for different sparsity regimes.}
                $k^{*}$ is the average number of concepts predicted by~\splice.
            }\label{fig:crm}
        \end{minipage}
         &
        \begin{minipage}[t]{\linewidth}
            \captionof{table}{
                \textbf{Evaluation of the concepts disentanglement with DCI metrics~\cite{eastwood2018a}.}
                We use linear probing to predict the video scenario, \eg, \emph{cooking} or \emph{sports}, from just the reconstruction weights. 
            }\label{tab:disentanglement}
        \end{minipage}
    \end{tabular}
    \vspace{-4mm}
\end{table*}

%% file: sections/figures/vocabulary.tex
\begin{table*}[t]
    \centering
    \caption{
        \textbf{Analysis of the concept vocabulary.} Few well-selected concepts are sufficient to retain most of the reconstruction quality of the more redundant larger vocabulary (left). The GPT-based vocabulary achieves good reconstruction despite minimal human effort in concepts definition (right).
    }\label{tab:vocabulary}
    \vspace{-2mm}
    \begin{subtable}[t]{0.49\textwidth}
        \centering
        \scriptsize
        \setlength{\tabcolsep}{4pt}
        \begin{tabularx}{\linewidth}{Xccc}
            \toprule
            \textbf{Vocabulary} & \textbf{Size} & \textbf{ConVisBench} ($\rho$ / $\tau$) & \textbf{CRM@50} \\
            \midrule
            Full Voc.           & 2139          & 56.3 / 40.1                            & 0.73            \\
            Pruned ($\tau=0.5$) & 79            & 42.1 / 29.5                            & 1.00            \\
            Pruned ($\tau=0.9$) & 343           & 51.4 / 36.4                            & 0.99            \\
            \bottomrule
        \end{tabularx}
    \end{subtable}
    \hfill
    \begin{subtable}[t]{0.49\textwidth}
        \centering
        \scriptsize
        \begin{tabularx}{\linewidth}{Xccc}
            \toprule
            \textbf{Vocabulary} & \textbf{Size} & \textbf{EgoMCQ} & \textbf{ConVisBench} ($\rho$ / $\tau$) \\
            \midrule
            EgoClip             & 700           & 35.43 / 81.35   & 51.0 / 36.2                            \\
            Full Voc.           & 2139          & 36.06 / 82.08   & 56.3 / 40.1                            \\
            GPT-5.4             & 414           & 34.76 / 80.28   & 52.3 / 37.1                            \\
            \bottomrule
        \end{tabularx}
    \end{subtable}
    \vspace{-4mm}
\end{table*}

%% file: sections/tables/image_retrieval_faithfullness.tex
\begin{table*}[t]
     \scriptsize
     \centering
     \caption{
          \textbf{Performance and Faithfulness on Image-to-Text (I$\rightarrow$T) and Text-to-Image (T$\rightarrow$I) retrieval tasks (Flickr30k and COCO)}. We report retrieval metrics and faithfulness.
     }
     \vspace{-2mm}
     \label{tab:image_retrieval}
     \setlength{\tabcolsep}{3.25pt}
     \begin{tabularx}{\linewidth}{Xcc|ccc|ccc|ccc|ccc}
          \toprule
          \multirow{2}{*}{\textbf{Method}}
           & \textbf{Concepts}                                      & \textbf{Zero}
           & \multicolumn{3}{c}{\textbf{Flickr30k I$\rightarrow$T}}
           & \multicolumn{3}{c}{\textbf{Flickr30k T$\rightarrow$I}}
           & \multicolumn{3}{c}{\textbf{COCO I$\rightarrow$T}}
           & \multicolumn{3}{c}{\textbf{COCO T$\rightarrow$I}}                                                                                                                       \\
           & \textbf{Vocab.}                                        & \textbf{Shot}     & R@1               & R@5               & R@10              & R@1               & R@5 & R@10
           & R@1                                                    & R@5               & R@10              & R@1               & R@5               & R@10                           \\
          \midrule
          \rowcolor{gray!50} \multicolumn{15}{c}{\emph{Task Metrics}}                                                                                                                \\
          \midrule
          CLIP (ViT-B)~\cite{radford2021learning}
           & N/A                                                    & \cmark
           & 67.17                                                  & 88.90             & 93.80             & 63.60             & 86.78             & 92.26
           & 51.90                                                  & 81.26             & 90.24             & 47.48             & 77.10             & 87.08                          \\
          \midrule
          SpLiCE~\cite{bhalla2024interpreting}
           & LAION                                                  & \cmark
           & 36.43                                                  & 65.03             & 75.36             & 36.83             & 63.50             & 74.20
           & 32.26                                                  & 62.94             & 76.44             & 29.92             & 59.22             & 73.00
          \\
          EXPLAIN-R~\cite{hoang-xuan2025advancing}
           & ImageNet                                               & \cmark
           & 45.67                                                  & 73.97             & 83.30             & \textbf{49.30}    & \textbf{78.45}    & \textbf{87.10}
           & \textbf{44.22}                                         & \textbf{75.28}    & \textbf{87.30}    & \textbf{47.42}    & \textbf{78.04}    & \textbf{88.12}
          \\
          \oursrow \textbf{\ours}
           & ImageNet                                               & \cmark
           & \underline{48.48}                                      & \underline{76.17} & \underline{84.50} & 42.84             & 69.97             & 79.26
           & 41.80                                                  & 71.98             & 83.66             & 33.56             & 63.34             & 75.92
          \\
          \oursrow \textbf{\ours} (\texttt{+}\emph{distill.})
           & ImageNet                                               & \cmark
           & \textbf{52.30}                                         & \textbf{79.30}    & \textbf{86.97}    & \underline{46.87} & \underline{74.43} & \underline{83.41}
           & \underline{42.78}                                      & \underline{72.84} & \underline{84.46} & \underline{36.34} & \underline{65.82} & \underline{78.66}
          \\
          \midrule
          EXPLAIN-R~\cite{hoang-xuan2025advancing}
           & Target                                                 & \xmark
           & 74.53                                                  & 92.58             & 96.24             & 73.60             & 92.67             & 96.30
           & 58.03                                                  & 85.68             & 93.19             & 56.30             & 84.98             & 93.09
          \\
          \midrule
          \rowcolor{gray!50} \multicolumn{15}{c}{\emph{Faithfulness}}                                                                                                                \\
          \midrule
          SpLiCE~\cite{bhalla2024interpreting}
           & LAION                                                  & \cmark
           & 40.25                                                  & 70.41             & 80.65             & 40.50             & 69.56             & 80.04
           & 38.12                                                  & 72.40             & 86.00             & 38.26             & 72.74             & 85.66
          \\
          EXPLAIN-R~\cite{hoang-xuan2025advancing}
           & ImageNet                                               & \cmark
           & 44.48                                                  & 72.62             & 81.42             & 42.52             & 70.02             & 79.44
           & 46.72                                                  & 77.82             & 88.32             & 39.68             & 70.10             & 83.52
          \\
          \oursrow \textbf{\ours}
           & ImageNet                                               & \cmark
           & \underline{55.24}                                      & \underline{84.27} & \underline{91.39} & \underline{51.77} & \underline{81.56} & \underline{89.64}
           & \underline{54.78}                                      & \underline{86.42} & \underline{94.38} & \underline{47.36} & \underline{81.60} & \underline{91.42}
          \\
          \oursrow \textbf{\ours} (\texttt{+}\emph{distill.})
           & ImageNet                                               & \cmark
           & \textbf{62.72}                                         & \textbf{90.19}    & \textbf{95.40}    & \textbf{55.94}    & \textbf{85.65}    & \textbf{92.85}
           & \textbf{58.16}                                         & \textbf{89.80}    & \textbf{96.22}    & \textbf{49.76}    & \textbf{84.24}    & \textbf{93.66}
          \\
          \midrule
          EXPLAIN-R~\cite{hoang-xuan2025advancing}
           & Target                                                 & \xmark
           & 66.37                                                  & 86.32             & 91.26             & 60.31             & 82.79             & 89.42
           & 61.99                                                  & 89.31             & 95.51             & 55.03             & 86.64             & 94.46
          \\
          \bottomrule
     \end{tabularx}
     \vspace{-4mm}
\end{table*}

%% file: sections/tables/image_cls_faithfullness.tex
\begin{table}[t]
    \scriptsize
    \centering
    \caption{
        \textbf{Accuracy and faithfulness on image classification.}
        For zero-shot evaluations, we train on ImageNet and evaluate the quality of the reconstructed embeddings on Flowers102 and~Food101.
    }
    \label{tab:image_classification}
    \vspace{0mm}
    \begin{tabularx}{\linewidth}{lcc*{6}{>{\centering\arraybackslash}X}}
        \toprule
        \multirow{2}{*}{\textbf{Method}}           & \textbf{Concepts} & \textbf{Zero} & \multicolumn{2}{c}{\textbf{ImageNet}} & \multicolumn{2}{c}{\textbf{Flowers102}} & \multicolumn{2}{c}{\textbf{Food101}}                                                             \\
                                                   & \textbf{Vocab.}   & \textbf{Shot} & Acc.                                  & Faith.                                  & Acc.                                 & Faith.            & Acc.              & Faith.            \\
        \midrule
        CLIP (ViT-B)~\cite{radford2021learning}    & N/A               & \cmark        & 61.92                                 & N/A                                     & 63.39                                & N/A               & 80.88             & N/A               \\
        \midrule
        SpLiCE~\cite{bhalla2024interpreting}       & LAION             & \cmark        & 42.91                                 & 51.33                                   & -                                    & 26.12             & -                 & 51.80             \\
        EXPLAIN-R~\cite{hoang-xuan2025advancing}   & ImageNet          & \cmark        & \textbf{50.77}                        & 50.19                                   & 15.51                                & 16.59             & 45.86             & 45.81             \\
        \oursrow \textbf{\ours}                             & ImageNet          & \cmark        & \underline{48.04}                     & \underline{60.15}                       & \textbf{32.14}                       & \textbf{41.78}    & \underline{62.67} & \underline{66.83} \\
        \oursrow \textbf{\ours} (\texttt{+}\emph{distill.}) & ImageNet          & \cmark        & 47.84                                 & \textbf{60.68}                          & \underline{28.82}                    & \underline{41.18} & \textbf{63.24}    & \textbf{67.49}    \\
        \midrule
        EXPLAIN-R~\cite{hoang-xuan2025advancing}   & Target-specific   & \xmark        & 52.37                                 & 53.38                                   & -                                    & 37.73             & -                 & 63.62             \\
        \bottomrule
    \end{tabularx}
    \vspace{-4mm}
\end{table}

%% file: sections/figures/shift_bias.tex
\begin{table*}[t]
    \centering
    \begin{tabular}{@{}p{0.49\textwidth}@{\hspace{0.02\textwidth}}p{0.49\textwidth}@{}}
        \begin{minipage}[t]{\linewidth}
            \vspace{0pt}
            \centering
            \includegraphics[width=0.9\linewidth]{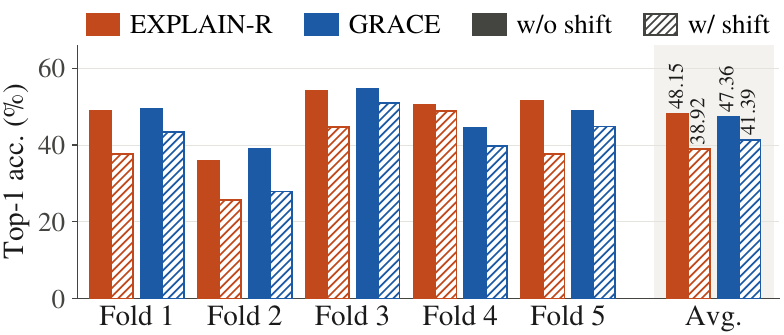}
        \end{minipage}
         &
        \begin{minipage}[t]{\linewidth}
            \vspace{0pt}
            \centering
            \scriptsize
            \vspace{2mm}
            \begin{tabular}{ll}
                \toprule
                \textbf{Label} & \textbf{Most activated Concepts}                                                                     \\
                \midrule
                volcano         & volcano, \textcolor{BrickRed}{\textbf{chile}}, view towards, \textcolor{BrickRed}{\textbf{indonesia}}, \dots \\
                rickshaw        & \textcolor{BrickRed}{\textbf{tokyo}}, flower delivery, bike, fast delivery, \dots \\
                cassette player & \textcolor{BrickRed}{\textbf{casio g}}, radio station, cassette, \dots \\
                digital clock   & clock, \textcolor{BrickRed}{\textbf{casio g}}, temperature, heartbeat, \dots           \\
                digital watch   & \textcolor{BrickRed}{\textbf{casio g}}, chronograph, clock, time, \dots              \\
                oscilloscope    & ultrasound, heartbeat, synthesis, \textcolor{BrickRed}{\textbf{casio g}}, \dots                 \\
                \bottomrule
            \end{tabular}
        \end{minipage} \\[-1mm]
        \begin{minipage}[t]{\linewidth}
            \captionof{figure}{
                \textbf{Transferability of the decomposition.}
                We train \ours on custom ImageNet folds with minimal concept overlap, demonstrating robust transfer under severe concept-distribution shifts.
            }\label{fig:shift}
        \end{minipage}
         &
        \begin{minipage}[t]{\linewidth}
            \captionof{table}{
                \textbf{Bias analysis of the decomposition.}
                We report the most activated concepts for some representative ImageNet classes, showing interesting patterns, \eg, strong \emph{entity-location} correlation.
            }\label{tab:bias_analysis}
        \end{minipage}
    \end{tabular}
    \vspace{-4mm}
\end{table*}

%% file: sections/figures/qualitatives.tex
\begin{figure*}[t]
    \centering
    \includegraphics[width=0.85\linewidth, trim={0 0 0 1.75cm},clip]{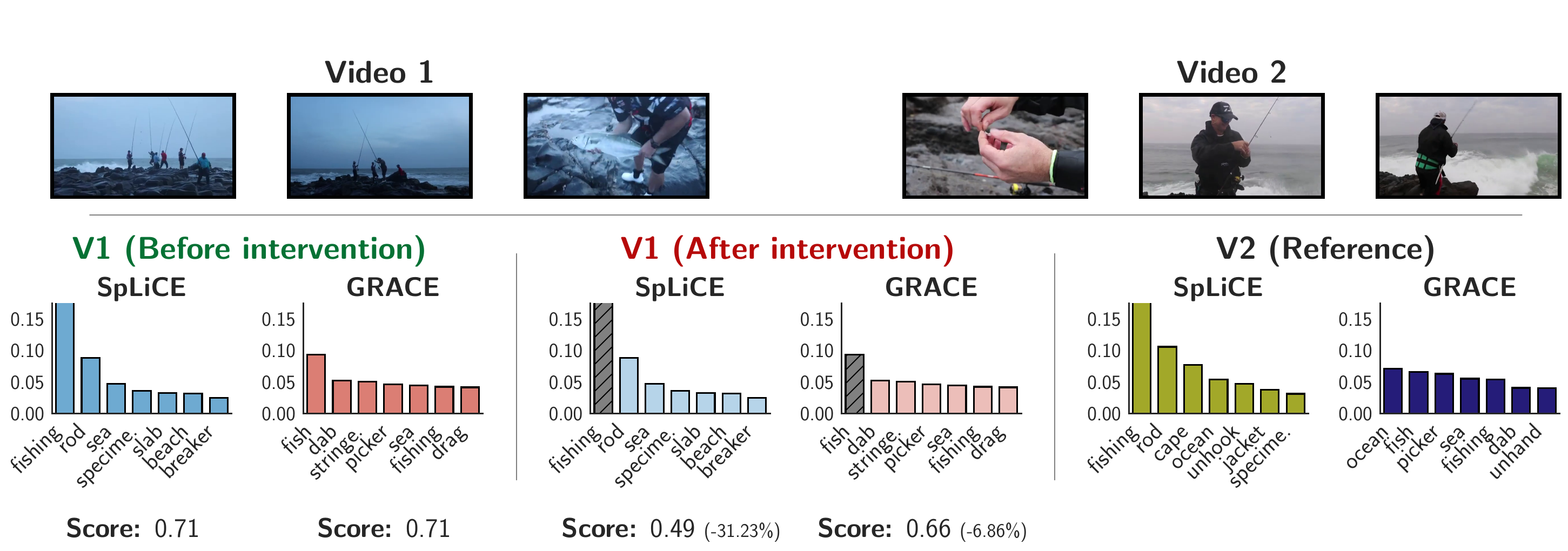}
    \vspace{-2mm}
    \caption{
        \textbf{Intervention on the reconstruction weights on ConVisBench~\cite{liberatori2025convisbench}.}
        We artificially zero out a concept from the predicted reconstruction weights and estimate the similarity \emph{before} and \emph{after} the intervention. 
        \splice focuses on a few concepts: when one of these is artificially removed, the similarity score drops significantly, even though the two videos remain similar in terms of subject, location, and objects. \ours captures the relevant concepts more uniformly.
    }
    \label{fig:qualitatives_interventions}
    \vspace{-4mm}
\end{figure*}

%% file: sections/conclusions.tex
\section{Conclusions}\label{sec:conclusions}
We explore how to make existing VLMs more interpretable.
We argue that these models already encode a latent concept-based structure, though it is not explicitly exposed.
Therefore, we introduce \ours, a novel framework that captures this structure to learn a general decomposition of visual embeddings that can generalize across tasks and domains.
Our approach overcomes many limitations of similar previous approaches, as it does not rely on per-sample optimization or external concept evidence. 
Instead, we exploit a reconstruction process that preserves the pairwise geometric relations among samples and encourages a selective and efficient use of the concept vocabulary.
We show that \ours decomposes visual embeddings into highly semantic concepts, which are well aligned with human judgement.
The decomposition transfers well to unseen domains, is robust to severe shift in the concept distribution, and can be used to debug dataset biases.

%% file: appendix/main.tex
\section*{Appendix}

This appendix is organized as follows:
\begin{itemize}
    \item \textbf{Implementation details}: In Sec.~\ref{sec:appendix_implementation_details}, we provide additional implementation details about \ours and the adaptation of previous works to our setting.
    \item \textbf{Disentanglement in reconstruction}: In Sec.~\ref{sec:appendix_disentanglement}, we present a more complete description of the metrics used to evaluate concepts disentanglement in \ours.
    \item \textbf{Concepts redundancy}: In Sec.~\ref{sec:appendix_redundancy}, we evaluate the redundancy in the concept decompositions produced by our approach.
    \item \textbf{Visual grounding}: In Sec.~\ref{sec:appendix_grounding}, we replace the image backbone with TIPSv2 end evaluate spatial grounding of the concept decomposition.
    \item \textbf{Captions integration in \ours}: In Sec.~\ref{sec:appendix_captions}, we describe the integration of a multi-modal distillation loss that can improve the multi-modal alignment of our method when captions are available during training.
    \item \textbf{Concepts projection in the reconstruction}: In Sec.~\ref{sec:appendix_projection}, we analyze how \ours encourages semantic alignment of the predicted reconstruction weights with human perception.
    \item \textbf{Robustness to the choice of the concept vocabulary}: In Sec.~\ref{sec:appendix_vocabulary}, we compare the use of different concept vocabularies in image and video benchmarks, showing the robustness of our method to different concepts sources.
    \item \textbf{Visual evidence and captions-based evidence}: In Sec.~\ref{sec:appendix_visual_concepts_evidence}, we compare two strategies to identify concepts in a visual input, \ie, \emph{visual evidence}, based on multi-modal comparison between the input and the embeddings of a given concept vocabulary, as in \ours, and \emph{captions-based evidence}, which leverages external captioning models to identify concepts.
    \item \textbf{Additional qualitative results}: In Sec.~\ref{sec:appendix_qualititatives}, we present more qualitative results showing the reconstruction weights predicted by our method and its robustness to concepts interventions.
    \item \textbf{Limitations and future works}: In Sec.~\ref{sec:appendix_limitations}, we describe the main limitations of this work and propose some possible future research directions. 
\end{itemize}

\input{appendix/sections/implementation_details}

\input{appendix/sections/disentanglement}
\input{appendix/sections/redundancy}
\input{appendix/sections/vis_grounding}
\input{appendix/sections/captions_integration}
\input{appendix/sections/ablations}
\input{appendix/sections/visual_concept_evidence}
\input{appendix/sections/qualitatives}
\input{appendix/sections/limitations.tex}

%% file: appendix/sections/implementation_details.tex
\section{Implementation details}\label{sec:appendix_implementation_details}

\subsection{Training and evaluation protocol}
For video benchmarks, we train \ours for 5 epochs on EgoClip~\cite{lin2022egocentric}, using a batch size of 128 and a learning rate of $0.0001$.
For image benchmarks, we train \ours for 25 epochs on ImageNet~\cite{deng2009imagenet}, using batch size 512 and learning rate $0.0001$.
During training, we select the model with the lowest reconstruction loss ($\mathcal{L}_{rec}$) on the training set.
For experiments that incorporate captions, we select the best model according to the faithfulness of the vision-language alignment on the training set.
We use $\tau=0.05$ for the temperature parameter in the reconstruction equation (Eq.~\ref{eq:reconstruction}).
We train \ours on NVIDIA RTX 4090 GPUs. 

\subsection{Video benchmarks}
We train \ours on the large-scale Ego4D dataset~\cite{grauman2022ego4d}, covering a large range of different human daily activities over 3.6k recording hours.
Specifically, we use the EgoClip~\cite{lin2022egocentric} subset, which consists of 3.8M short video clips paired with human-annotated captions.
Videos are annotated with a \emph{scenario} label that identifies the main activities appearing in the video, such as \emph{cooking} or \emph{playing sports}.
For the consistency loss $\mathcal{L}_{con}$, we use the scenario label to group together samples we expect to activate similar concepts.
Then, we evaluate \ours on different benchmarks in zero-shot to test the multi-modal alignment of the reconstructed visual embeddings and the quality and interpretability of the reconstruction weights.

For video benchmarks, we use two concept vocabularies: a \emph{small} vocabulary, built from verbs and nouns in EgoClip captions grouped via the official fixed taxonomy provided by the authors (700 concepts), and a \emph{large} vocabulary, consisting of the top 25\% most frequent verbs and nouns across all captions (approx. 2.1K concepts).
In all experiments, we use the same vocabulary at both training and inference time.

\textbf{EgoMCQ}~\cite{lin2022egocentric} is a multiple-choice text-to-video retrieval benchmark, designed to evaluate video-text alignment. 
It consists of 39K questions, each comprising a textual query and five candidate video clips, only one of which is the correct match.
The benchmark includes two evaluation splits: (i) an \emph{Inter} split, where candidate clips are drawn from different videos, and (ii) an \emph{Intra} split, where all candidates are sampled from the same video.
For retrieval, we compute cosine similarity between the text embedding of the query and the reconstructed visual embeddings from \ours. Performance is evaluated using~accuracy.

\textbf{Charades-Ego}~\cite{sigurdsson2018charades} is a large-scale dataset of first and third person recordings, covering around 68 hours of daily human activities. We evaluate our method on the test split of the Action Recognition benchmark, covering 846 short videos annotated with one or more activity labels from a taxonomy of 157 classes.
For this task, we define a zero-shot classifier by extracting the text embeddings of the class labels and computing their similarity with respect to the reconstructed visual embeddings.
Performance is measured with mean Average Precision~(mAP).

\textbf{ConVisBench}~\cite{liberatori2025convisbench} is a dataset consisting of 610 video pairs sampled from 543 different videos. 
Each pair is annotated according to human-judged similarity across five different macro categories: \emph{Main Action}, \emph{Main Subjects}, \emph{Main Objects}, \emph{Location} and \emph{Order of Actions}. 
Videos are also annotated with short captions.
Similarity is collected by showing the video pairs to different annotators and asking them to evaluate the similarity of the two videos on a 1-to-5 scale.
We evaluate \ours on the \emph{Concept-based Video Similarity Estimation} task, where the goal is to predict a similarity score based on the semantic similarity between the two videos, which is compared with human judgment with the Spearman’s rank correlation $\rho$ and Kendall’s $\tau$. 
These metrics evaluate how the predictions correlate with the similarity scores collected from human annotators. 
We compute the similarity score for a video pair using the CLIPScore~\cite{hessel2021clipscore}, \ie, we compute the cosine similarity between the reconstructed visual embedding of the first video and the text embedding of the second video’s caption, and vice versa, and then average the two scores.
Moreover, we also compute the distance between the reconstruction weights of the visual embeddings of the two videos, as predicted by \ours.
This approach allows us to directly evaluate the expressivity of the reconstruction weights, and how they align with human interpretable semantic similarities or differences between the two~videos.

\subsubsection{Adaptation of \splice and \explainr to video benchmarks}
\splice~\cite{bhalla2024interpreting} and \explainr~\cite{hoang-xuan2025advancing} were originally proposed for image benchmarks, and we adapt them to work on video benchmarks with some changes.
For \textbf{\splice}, we tune the $\ell_1$ penalty parameter $\lambda$ to achieve a reasonable average sparsity in the reconstruction weights. 
Specifically, we choose $\lambda=0.1$ when using the \emph{small} concepts vocabulary on EgoMCQ and Charades, which results in approximately 42.2 and 37.9 non-zero weights on the two datasets, respectively. 
On ConVisBench, we set $\lambda=0.15$.
For \textbf{\explainr}, we uniformly sample 8 frames from each video clip and extract frame-level captions using BLIP2~\cite{li2023blip}. 
This process takes approximately 3k GPU hours on EgoClip. 
Following the original protocol, we then compute the concept encoding by extracting lemmas that appear in at least three captions and filtering out those that occur in more than 10\% of the total captions, obtaining a vocabulary of 10990~concepts.

\subsection{Image benchmarks}
We train \ours on ImageNet~\cite{deng2009imagenet} with a CLIP ViT-B/32 backbone~\cite{radford2021learning}. 
Following the protocol introduced in EXPLAIN-R~\cite{hoang-xuan2025advancing}, we generate 10 captions per image using BLIP2~\cite{li2023blip} and build the concept vocabulary by filtering out both rare and overly frequent concepts extracted from the captions. 
For $\mathcal{L}_{con}$, we use the image labels as category label, encouraging the model to learn similar reconstruction weights for visual samples belonging to the same class.
The visual and text encoders are kept frozen, and we pre-extract the visual embeddings before training to significantly speed up the training process.

We evaluate \ours on different Image Classification and Image-Text Retrieval tasks without additional fine-tuning, following the evaluation splits defined by EXPLAIN-R.
For Image Classification, we consider \textbf{ImageNet}~\cite{deng2009imagenet} (validation split), \textbf{Food101}~\cite{food101} (test split) and \textbf{Flowers102}~\cite{flowers102} (test split).
For these tasks, we define a zero-shot classifier using the ground truth labels of each dataset, and compare their text embeddings with the reconstructed visual embeddings using cosine similarity.
For Image-Text Retrieval, we include \textbf{Flickr30k}~\cite{plummer2015flickr30k} (entire dataset) and \textbf{COCO}~\cite{lin2014microsoft} (validation~set).

%% file: appendix/sections/disentanglement.tex
\section{Disentanglement in the reconstruction}\label{sec:appendix_disentanglement}
In this section, we evaluate the impact of the consistency regularization described in Sec.~\ref{sec:method_regularizations}.
The regularization term enforces consistency in the reconstruction weights for samples sharing the same category, and penalizes similarity when samples belong to different categories.
We evaluate the effectiveness of this approach in disentangling the concepts that contribute the most in each category by following the protocol defined in \cite{eastwood2018a,eastwood2023dcies}, which we summarize in the following. 

\textbf{DCI Metrics.} In this framework, a set of sparse regressors $f$ is trained on the target representations $\mathbf{c} \in \mathbb{R}^D$ to predict the latent factors $\mathbf{z}\in \mathbb{R}^K$, and evaluate disentanglement with respect to these predicted factors. $\mathbf{P} \in \mathbb{R}^{D \times K}$ represents the relative importance of each code $\mathbf{c}_i$ in predicting a given latent factor $\mathbf{z}_j$.
The classifiers are then evaluated in terms of \emph{informativeness}, \emph{disentanglement} and \emph{completeness}.
\textbf{Informativeness} is the accuracy of the regressor in predicting the ground truth latent factors on a held-out test set.
\textbf{Disentanglement} measures how close each code $\mathbf{c}_i$ is to capturing one single latent factor, and is defined as $D_i = (1 - H_K (P_{i\cdot}))$, where $H_K (P_{i\cdot})$ is the entropy of the relative importance of code $i$ over all the latent factors. In the ideal case, each code contributes to predicting a single latent factor, hence resulting in $D_i = 1$.
\textbf{Completeness} measures whether each latent factor is captured by one or more codes, and is defined as $C_j = (1 - H_D (P_{\cdot j}))$. When a latent factor is captured by a single code, the corresponding completeness is~$C_j = 1$.

We adapt these metrics to our setting by evaluating the quality of the reconstruction weights from \ours (\emph{codes}) with respect to some ground truth category label (\emph{latent factors}). Specifically, we consider the \emph{scenario} annotations in Ego4D, a set of discrete labels that describe the general context of a video in the dataset, as category labels for computing the metrics.
Since these labels are binary, rather than continuous latent factors, we use logistic regression classifiers with $\ell_1$ penalties to predict the category labels from the reconstruction weights and define $\mathbf{P}_{ij}$ as the weight assigned by classifier $j$ to the reconstruction weight $i$, normalized with respect to all the classifiers. For \emph{informativeness}, we compute the mean average precision on the category predictions.
From EgoMCQ, we derive a subset of videos from the 25 most common scenarios, and divide them into 27453 and 9206 training and validation samples.

%% file: appendix/sections/redundancy.tex
\section{Concepts redundancy}\label{sec:appendix_redundancy}

To analyze redundancy within the concept-based decomposition, we cluster the LAION vocabulary concepts using K-Means applied to their text embeddings and select the optimal number of clusters via the silhouette score. This produces 120 distinct clusters with a median size of 110.5 samples. We present representative examples of the resulting macro-clusters in Table~\ref{tab:redundancy}-left, and assign each macro-cluster the label closest to the centroid. Each cluster contains synonyms and related words.
\begin{table}[t]
    \centering
    \caption{
        \textbf{Analysis of concepts redundancy.} Representative macro-clusters obtained with K-Means on the text embeddings of the LAION concept vocabulary (left). Median number of distinct macro-clusters among the top-25 concepts per class, with and without the calibration module (right).
    }\label{tab:redundancy}
    \begin{subtable}[t]{0.64\linewidth}
        \centering
        \scriptsize
        \begin{tabularx}{\linewidth}{l c X}
            \toprule
            \textbf{Macro cluster} & \textbf{Size} & \textbf{Concepts}                                                                                 \\
            \midrule
            Kitchen                & 83            & \textit{kitchen}, \textit{kitchen design}, \textit{small kitchen}, \textit{kitchen pantry}, \dots \\
            Staring                & 101           & \textit{staring}, \textit{gazing}, \textit{focused}, \textit{observing}, \textit{admiring}, \dots \\
            Glass window           & 74            & \textit{glass window}, \textit{glass door}, \textit{window}, \textit{old window}, \dots            \\
            \bottomrule
        \end{tabularx}
    \end{subtable}
    \hfill
    \begin{subtable}[t]{0.33\linewidth}
        \centering
        \scriptsize
        \begin{tabular}{l c}
            \toprule
            \textbf{Model}        & \textbf{\# Clusters} \\
            \midrule
            \ours                 & \textbf{16.0}        \\
            \ours w/o calibration & 9.0                  \\
            \bottomrule
        \end{tabular}
    \end{subtable}
\end{table}
Following clustering, we take the top-25 most activated concepts for each class and map them to their corresponding macro-clusters. This results in a median of 16.0 distinct clusters among the top-25 concepts per class, indicating relatively low redundancy.
The calibration module controls the contribution of each concept in the reconstruction and is crucial for reducing redundancies. Indeed, when this module is removed the median number of distinct macro concepts in Top-25 drops to 9.0, showing significantly higher redundancy (Table~\ref{tab:redundancy}-right).

%% file: appendix/sections/vis_grounding.tex
\section{Visual grounding}\label{sec:appendix_grounding}

\begin{wraptable}{r}{0.32\linewidth}
    \vspace{-1.2em}
    \centering
    \small
    \caption{\textbf{Concept grounding on ADE20k.} We report AUPRC.}
    \label{tab:appendix_grounding}
    \begin{tabular}{lc}
        \toprule
        Method & AUPRC \\
        \midrule
        BLIP-2~\cite{li2023blip}                 & 0.098          \\
        LLaVA-NeXT~\cite{li2024llava}            & 0.302          \\
        ConceptScope~\cite{choi2026conceptscope} & 0.399          \\
        \oursrow \textbf{\ours (TIPSv2)} & \textbf{0.482} \\
        \bottomrule
    \end{tabular}
    \vspace{-1em}
\end{wraptable}

\ours performs concept-based decomposition of the embedding corresponding to the CLS token, leveraging its alignment with text. 
In CLIP and PerceptionEncoder, the embeddings corresponding to spatial visual patches are not text-aligned, so this decomposition cannot be applied to them directly.

To address this, we extend \ours to a different VLM backbone, TIPSv2~\cite{cao2026tipsv}, which outputs text-aligned visual patch embeddings that can be decomposed with \ours. 
We train \ours with TIPSv2 on ImageNet, using the original reconstruction objective on the CLS token. 
At inference time, we apply the same concept-based decomposition to the spatial patches and evaluate spatial grounding on ADE20k~\cite{zhou2017scene}, selecting the best class-concept mapping on the training set by following the evaluation protocol from ConceptScope~\cite{choi2026conceptscope}.

\begin{figure}
    \centering
    \includegraphics[width=0.8\linewidth]{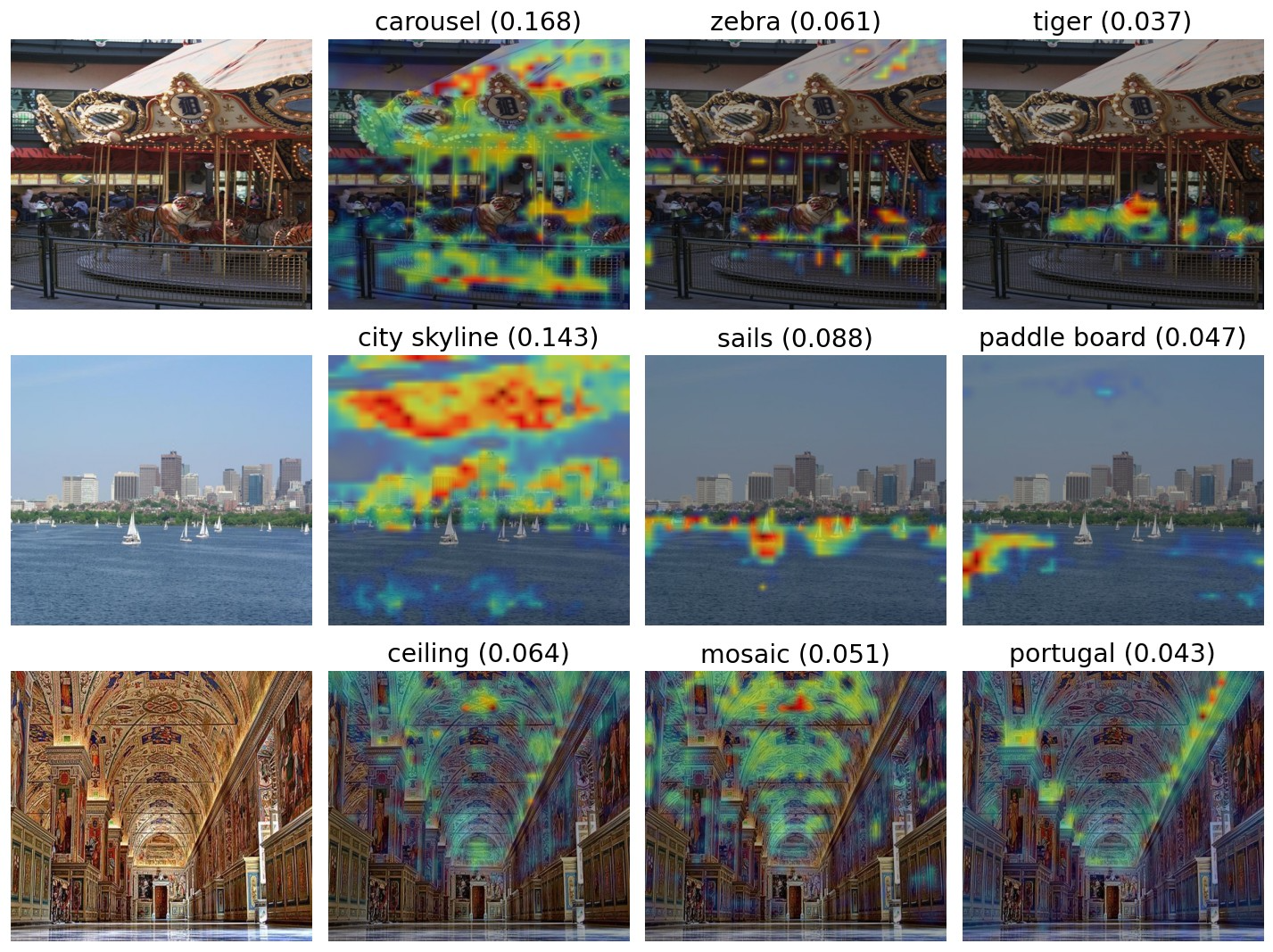}
    \caption{\textbf{Visual grounding of the concept decomposition on ADE20k.}\label{fig:ade20k_viz}
    Using a \ours variant trained on TIPSv2~\cite{cao2026tipsv}, we compute the concept decomposition of the CLS token, select the top three highest-scoring concepts, and show their contribution in the visual patches.}
\end{figure}

As shown in Table~\ref{tab:appendix_grounding}, \ours substantially improves concept grounding compared to ConceptScope (0.482 vs 0.399), showing that concepts can be reliably localized in the input image. 
These experiments further show that \ours can be easily implemented on top of different backbones, including CLIP, PerceptionEncoder, and TIPSv2, for both global and dense concept-based decomposition tasks.
In Fig.~\ref{fig:ade20k_viz}, we select a few samples from ADE20k, compute the concept decomposition of the CLS token, select the top three highest-scoring concepts, and show their contribution in the visual patches.

%% file: appendix/sections/captions_integration.tex
\section{Captions integration in \ours}\label{sec:appendix_captions}
The reconstructed visual embeddings from \ours achieve high faithfulness with respect to the original embeddings and strong downstream task performance, despite not relying on instance-level optimization or captions during training or inference.
For a fair comparison with EXPLAIN-R, which uses captions to identify concepts in the visual input and to produce surrogate representations that preserve the vision-language alignment of the target model, we also consider a variant of \ours that leverages captions at training time to further improve the reconstruction process.
Specifically, we incorporate the distillation loss from~\cite{hoang-xuan2025advancing} to push the reconstructed embeddings to match the original embeddings in terms of cosine similarity with the text embeddings of the captions in the batch.
For any visual $\mathbf{z}_i$ and text $\mathbf{t}_j$ embeddings in the batch $\mathcal{B}$, we indicate the distribution of the pairwise image-to-text cosine similarities as:
\begin{equation}
    P_i = \frac{\exp(\mathbf{z}_i^T\mathbf{t}_i / \tau)}{\sum_k \exp(\mathbf{z}_i^T\mathbf{t}_k / \tau)},
\end{equation}
where $\tau$ is a temperature parameter, and the same distribution for the reconstructed embeddings $\mathbf{\tilde{z}}_i$ as:
\begin{equation}
    \tilde{P}_i = \frac{\exp(\mathbf{\tilde{z}}_i^T\mathbf{t}_i / \tau)}{\sum_k \exp(\mathbf{\tilde{z}}_i^T\mathbf{t}_k / \tau)}.
\end{equation}
Similarly, we indicate with $Q_i$ and $\tilde{Q}_i$ the distributions of the text-to-image similarities.
The distillation loss $\mathcal{L}_{d}$ is defined as the KL divergence between the distributions obtained from the original and reconstructed embeddings:
\begin{equation}
    \mathcal{L}_{d} = \frac{1}{2|\mathcal{B}|} \left[ \sum_{i=1}^{|\mathcal{B}|}\text{KL}(P_i||\tilde{P}_i) + \sum_{j=1}^{|\mathcal{B}|}\text{KL}(Q_j||\tilde{Q}_j)\right].
\end{equation}
In our experiments, we observe that the distillation provides a quite marginal contribution on both image and video benchmarks, indicating that uni-modal reconstruction with our geometry-aware objectives alone is sufficient to produce faithful representations with respect to the original embeddings.

%% file: appendix/sections/ablations.tex
\input{appendix/tables/concepts_projection}

\section{Concepts projection in the reconstruction}\label{sec:appendix_projection}
Visual embeddings in \ours are reconstructed from a linear combination of the text embeddings of the concepts in the vocabulary.
In the reconstruction process, concepts are projected to facilitate the reconstruction (Eq.~\ref{eq:reconstruction}).
Ideally, this process should limit concept entanglement, which would limit the interpretability of the reconstruction weights, thus preserving the pairwise relations between the concepts in the projected space.
In Table~\ref{tab:appendix_concepts_proj}, we compare different strategies in terms of constraints and capacity, starting from a simple identity mapping and evaluating the quality of the reconstruction in terms of multi-modal alignment on EgoMCQ~\cite{lin2022egocentric} and semantic alignment of the reconstruction weights on ConVisBench~\cite{liberatori2025convisbench}.

The identity mapping achieves the best performance on ConVisBench: freezing the concept embeddings encourages the model to predict highly semantic weights to meet the reconstruction objectives. 
However, the performance and faithfulness on EgoMCQ are quite poor, as the identity is not sufficient to reconstruct the geometric relations of the original samples.
Replacing the identity with a linear projection or an MLP significantly improves multi-modal alignment and faithfulness of the reconstructed visual embeddings, at the cost of a lower interpretability of the reconstruction weights, as they become less correlated with human-judged similarity on ConVisBench.
The orthogonal linear projection in \ours strikes a good balance between the two aspects, as it achieves close to optimal multi-modal alignment and consistently better correlation on ConVisBench.
During training, we force $\mathbf{Q}$ to be orthogonal through a Cayley transform.

\input{appendix/tables/image_vocabulary}

\input{appendix/tables/evidence}

\section{Robustness to the choice of the concept vocabulary}\label{sec:appendix_vocabulary}

We present strong empirical evidence that \ours is robust to the choice of the concept vocabulary.

\textbf{ImageNet vs. LAION vocabularies.} For image benchmarks, we compare vocabularies derived from ImageNet and LAION captions. The ImageNet vocabulary was introduced in the EXPLAIN-R paper~\cite{hoang-xuan2025advancing}, mined from a large corpus of ImageNet captions extracted with BLIPv2, while the LAION vocabulary is built from bigrams extracted from LAION captions. The ImageNet and LAION vocabularies include 18490 and 15000 concepts, respectively. 
We compare the impact of using different concepts vocabularies in \ours on Image Retrieval (Table~\ref{tab:appendix_image_voc_retrieval}) and Image Classification tasks (Table~\ref{tab:appendix_image_voc_cls}). 
Task metrics and faithfulness are quite similar across all benchmarks, showing the robustness of \ours to different concept vocabularies.

\textbf{Video benchmarks.} For video benchmarks, we draw concepts from human-annotated captions at two levels of granularity: ``EgoClip'' and ``Full Voc.''. ``EgoClip'' uses the official Ego4D taxonomy of verbs and nouns~\cite{lin2022egocentric}, with related terms grouped into broader macro-concepts (e.g., ``brush'', ``clean'', ``dust'', ``mop'', ``sanitize'', and ``scrub'' are merged into a single ``clean'' concept). ``Full Voc.'' instead keeps all verbs and nouns from the captions ungrouped. These vocabularies contain 700 and 2139 concepts, respectively. We also test a vocabulary generated by prompting GPT5.4 to produce an exhaustive list of plausible text-based concepts likely to appear in human activity videos. These comparisons are reported in Table~\ref{tab:vocabulary}-right. We find that ConVisBench benefits considerably from richer, more exhaustive vocabularies, since they allow for finer-grained concept decomposition, whereas retrieval performance (EgoMCQ) is largely less affected by this choice.

\textbf{Concept filtering.} Using Singular Value Decomposition (SVD) on the concept matrix, we estimate the redundancy present in the concept vocabulary and show that it can be substantially compressed into fewer concepts without significantly hurting the downstream performance (Table~\ref{tab:vocabulary}-left).

Together, these experiments validate GRACE's robustness along three axes: the source of the concept vocabulary (ImageNet vs. LAION, human-annotated captions vs. GPT), its granularity (``EgoClip'' vs. ``Full Voc.'' for video benchmarks), and the impact of aggressive filtering (concept pruning).

%% file: appendix/tables/concepts_projection.tex
\begin{table}
    \scriptsize
    \centering
    \setlength{\tabcolsep}{3.25pt}
    \caption{
        \textbf{Concept projection in the reconstruction.} 
        We consider different strategies to project the concept embeddings before reconstruction, and evaluate retrieval accuracy on EgoMCQ~\cite{lin2022egocentric} and semantic alignment of the reconstruction weights on ConVisBench~\cite{liberatori2025convisbench}.
    }\label{tab:appendix_concepts_proj}
    \begin{tabular}{lcccc}
        \toprule
        \multirow{2}{*}{\textbf{Projection}} & \multicolumn{3}{c}{\textbf{EgoMCQ}} & \textbf{ConVisBench} \\
        & Intra Acc. & Inter Acc. & Faith. & \multicolumn{1}{c}{$\rho$ / $\tau$} \\
        \midrule
        Identity                             & 32.59             & 75.93             & 67.08             & \textbf{53.9} / \textbf{38.4} \\
        Linear                               & \underline{36.08} & \underline{81.33} & \underline{79.94} & 48.3 / 34.2 \\
        MLP                                  & \textbf{36.15}    & 81.12             & \textbf{80.37}    & 47.9 / 33.9 \\
        \oursrow \textbf{Linear (Orthogonal) - \ours} & 35.43             & \textbf{81.35}    & 77.86             & \underline{51.0} / \underline{36.2} \\
        \bottomrule
    \end{tabular}
\end{table}

%% file: appendix/tables/image_vocabulary.tex
\begin{table}[t]
    \scriptsize
    \centering
    \caption{\textbf{Comparison of different concept vocabularies on Image Classification tasks.} We compare different versions of \ours using the ImageNet~\cite{deng2009imagenet} and LAION~\cite{schuhmann2022laion} vocabularies, showing quite consistent performance across different vocabularies.}
    \label{tab:appendix_image_voc_cls}
    \begin{tabular}{lcccccccc}
        \toprule
        \multirow{2}{*}{\textbf{Method}} & \multirow{2}{*}{\textbf{Captions}}
        & \textbf{Concepts} 
        & \multicolumn{2}{c}{\textbf{ImageNet}} 
        & \multicolumn{2}{c}{\textbf{Flowers102}} 
        & \multicolumn{2}{c}{\textbf{Food101}} \\
        & & \textbf{Vocab.}   
        & Acc. & Faith.
        & Acc. & Faith. 
        & Acc. & Faith. \\
        \midrule
        %
        \ours           & \xmark                & ImageNet
                        & \underline{48.04}     & 60.15
                        & \textbf{32.14}        & \textbf{41.78}  
                        & \underline{62.67}     & \underline{66.83}  \\
        %
        \ours           & \xmark                & LAION
                        & 47.36                 & 59.87
                        & 25.11                 & 31.92
                        & 61.60                 & 64.84  \\
        \midrule
        %
        \ours           & \cmark                & ImageNet
                        & 47.84                 & \underline{60.68}
                        & \underline{28.82}     & \underline{41.18}
                        & \textbf{63.24}        & \textbf{67.49}  \\
        %
        \ours           & \cmark                & LAION
                        & \textbf{48.27}        & \textbf{60.96}
                        & 28.57                 & 36.79  
                        & 60.80                 & 64.65  \\
        \bottomrule
    \end{tabular}
\end{table}

\begin{table*}[t]
    \scriptsize
    \centering
    \setlength{\tabcolsep}{4pt}
    \caption{\textbf{Comparison of different concept vocabularies on Image-Text Retrieval tasks.} We compare \ours using the ImageNet~\cite{deng2009imagenet} and LAION~\cite{schuhmann2022laion} vocabularies, showing the robustness of our method to different vocabularies.}
    \label{tab:appendix_image_voc_retrieval}
    \begin{tabular}{lcc|ccc|ccc|ccc|ccc}
        \toprule
        \multirow{2}{*}{\textbf{Method}} & \multirow{2}{*}{\textbf{Captions}}
                                         & \textbf{Concepts}
                                         & \multicolumn{3}{c}{\textbf{Flickr30k I$\rightarrow$T}}
                                         & \multicolumn{3}{c}{\textbf{Flickr30k T$\rightarrow$I}}
                                         & \multicolumn{3}{c}{\textbf{COCO I$\rightarrow$T}}
                                         & \multicolumn{3}{c}{\textbf{COCO T$\rightarrow$I}}                                                     \\
                                       & & \textbf{Vocab.}                                        & R@1   & R@5   & R@10  & R@1   & R@5   & R@10
                                         & R@1                                                    & R@5   & R@10  & R@1   & R@5   & R@10         \\
        \midrule
        \rowcolor{gray!50} \multicolumn{15}{c}{\emph{Task Metrics}}                                                                          \\
        \midrule
        %
        \ours                            & \xmark & ImageNet
                                         & 48.48                                                  & 76.17 & 84.50 & 42.84 & 69.97 & 79.26
                                         & 41.80                                                  & 71.98 & 83.66 & 33.56 & 63.34 & 75.92        \\
        %
        \ours                            & \xmark & LAION
                                         & 48.24                                                  & 75.85 & 84.23 & 42.49 & 69.55 & 78.88
                                         & 41.32                                                  & 71.88 & 83.70 & 32.36 & 61.54 & 74.18        \\
        \ours                            & \cmark & ImageNet
                                         & \underline{52.30}    & \textbf{79.30}    & \underline{86.97} & \underline{46.87} & \underline{74.43} & \underline{83.41}
                                         & \underline{42.78}    & \underline{72.84} & \underline{84.46} & \textbf{36.34}    & \underline{65.82} & \underline{78.66}        \\
        %
        \ours                            & \cmark & LAION
                                         & \textbf{52.56}       & 79.29             & \textbf{87.18}    & \textbf{47.24}    & \textbf{74.62} & \textbf{83.66}
                                         & \textbf{43.48}       & \textbf{73.88}    & \textbf{84.80}    & \underline{35.94} & \textbf{66.60} & \textbf{79.18}        \\
        \midrule
        \rowcolor{gray!50} \multicolumn{15}{c}{\emph{Faithfulness Metrics}}                                                                              \\
        \midrule
        %
        \ours                            & \xmark & ImageNet
                                         & 55.24                                                  & 84.27 & 91.39 & 51.77 & 81.56 & 89.64
                                         & 54.78                                                  & 86.42 & 94.38 & 47.36 & 81.60 & 91.42        \\
        %
        \ours                            & \xmark & LAION
                                         & 55.19                                                  & 83.99 & 91.13 & 51.89 & 81.82 & 89.91
                                         & 54.38                                                  & 87.08 & 94.84 & 46.82 & 81.44 & 91.14        \\
        %
        \ours                            & \cmark & ImageNet
                                         & \underline{62.72}    & \underline{90.19} & \underline{95.40} & \underline{55.94} & \underline{85.65} & \underline{92.85}
                                         & \underline{58.16}    & \underline{89.80} & \underline{96.22} & \underline{49.76} & \underline{84.24} & \underline{93.66}        \\
        \ours                            & \cmark & LAION
                                         & \textbf{63.10}       & \textbf{90.45}    & \textbf{95.79}    & \textbf{56.69}    & \textbf{86.37}    & \textbf{93.35}
                                         & \textbf{58.56}       & \textbf{90.58}    & \textbf{96.88}    & \textbf{50.26}    & \textbf{85.44}    & \textbf{94.50}\\
        \bottomrule
    \end{tabular}
\end{table*}

%% file: appendix/tables/evidence.tex
\begin{table*}[tb]
    \scriptsize
    \centering
    \setlength{\tabcolsep}{3.5pt}
    \caption{\textbf{Comparison of visual evidence and captions-based evidence.} \emph{Visual evidence} identifies concepts in the input through the similarity of the visual embeddings and the text embeddings of the concepts in the vocabulary. \emph{Captions-based evidence} leverages an external captioning model to identify concepts in the input, which is the default approach for \explainr.}
    \label{tab:visual_captions_evidence}
    \begin{tabularx}{\linewidth}{Xccccccccccccccc}
        \toprule
        \multirow{2}{*}{\textbf{Method}} & \multirow{2}{*}{\textbf{Evidence}}
        & \multicolumn{2}{c}{\textbf{EgoMCQ}}
        & \multicolumn{2}{c}{\textbf{Charades}}
        & \multicolumn{2}{c}{\textbf{ImageNet}}
        & \multicolumn{2}{c}{\textbf{Flowers102}}
        & \multicolumn{2}{c}{\textbf{Food101}}
        & \multicolumn{2}{c}{\textbf{Flickr30k}}
        & \multicolumn{2}{c}{\textbf{COCO}} \\
        & 
        & Acc. & Faith.  
        & Acc. & Faith.  
        & Acc. & Faith.  
        & Acc. & Faith.  
        & Acc. & Faith.  
        & R@1 & Faith.   
        & R@1 & Faith.   
        \\
        \midrule
        %
        %
        %
        \explainr & Captions
            & 53.14             & 63.70             
            & 21.73             & 25.92             
            & \underline{50.77} & 50.19             
            & 15.51             & 16.59             
            & 45.86             & 45.81             
            & \underline{45.67} & 44.48             
            & \textbf{44.22}    & 46.72             
        \\
        %
        %
        %
        \explainr & Visual
            & \underline{57.15} & \underline{73.94} 
            & \underline{24.88} & \textbf{44.50}    
            & \textbf{52.51}    & \textbf{67.72}    
            & \underline{26.67} & \underline{35.00} 
            & \underline{55.65} & \underline{61.03} 
            & 37.78             & \underline{47.14} 
            & 35.16             & \underline{51.34} 
        \\
        \oursrow \textbf{\ours} & Visual 
            & \textbf{58.39}    & \textbf{77.86}    
            & \textbf{25.99}    & \underline{35.15} 
            & 48.04             & \underline{60.15} 
            & \textbf{32.14}    & \textbf{41.78}    
            & \textbf{62.67}    & \textbf{66.83}    
            & \textbf{48.48}    & \textbf{55.24}    
            & \underline{41.80} & \textbf{54.78}    
            \\
        \bottomrule
    \end{tabularx}
\end{table*}

%% file: appendix/sections/visual_concept_evidence.tex
\section{Visual evidence and captions-based evidence}\label{sec:appendix_visual_concepts_evidence}
\ours identifies concepts in the input by computing the cosine similarity between the visual embedding $\mathbf{z}_i$ of the input and the text embedding $\mathbf{c}_j$ of the concepts in the vocabulary, i.e., $\alpha_{i,j} = \cos(\mathbf{z}_i, \mathbf{c}_j)$. We refer to this signal as \emph{visual evidence}.
In contrast, \explainr relies on an external captioning model to generate $n$ captions for each input and constructs a concept encoding vector $\vect{\alpha}_i$, where $\alpha_{i,j}$ corresponds to the relative frequency of concept $j$ in the generated captions for that sample. We refer to this signal as \emph{caption-based evidence}.

Visual evidence does not depend on external models and introduces negligible overhead at inference time. 
More importantly, visual evidence provides a direct means to probe how semantic concepts are represented within the target model’s embeddings.
To further demonstrate the benefit of visual evidence, even beyond our method, we replace concept-based evidence in \explainr with visual evidence, while keeping the rest of the method unchanged.
Specifically, we define a concept encoding vector $\vect{\nu}_{i}$, such that $\nu_{i,j}$ is 1 if concept $j$ is among the top-k concepts matched by cosine similarity with the visual embedding, and 0 otherwise:
\begin{equation}
\nu_{i,j} = \begin{cases}
   \text{1} &\quad\text{if j} \in \text{TopK}_l \left[ \cos(\mathbf{z}_i, \mathbf{c}_l) \right]\\
   \text{0} &\quad\text{otherwise.} \\ 
 \end{cases}    
\end{equation}
We compare visual and captions-based evidence on video and image benchmarks in Table~\ref{tab:visual_captions_evidence}.
\explainr with visual evidence outperforms the original captions-based version on several benchmarks, in both task metrics and faithfulness.
These results demonstrate that visual evidence is sufficient to produce surrogate embeddings that preserve the performance of the original model.
\ours is designed to exploit and refine visual evidence, which results in consistent improvements across the benchmarks.

%% file: appendix/sections/qualitatives.tex
\section{Additional qualitative results}\label{sec:appendix_qualititatives}

\begin{figure*}[t]
    \centering
    \begin{subfigure}[b]{0.48\linewidth}
        \centering
        \includegraphics[width=\linewidth]{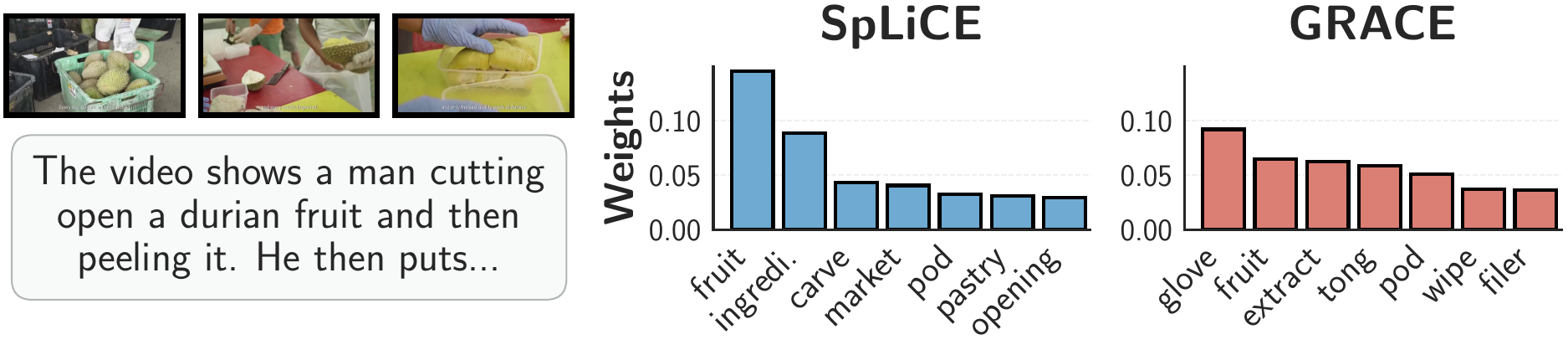}
    \end{subfigure}
    \hfill
    \begin{subfigure}[b]{0.48\linewidth}
        \centering
        \includegraphics[width=\linewidth]{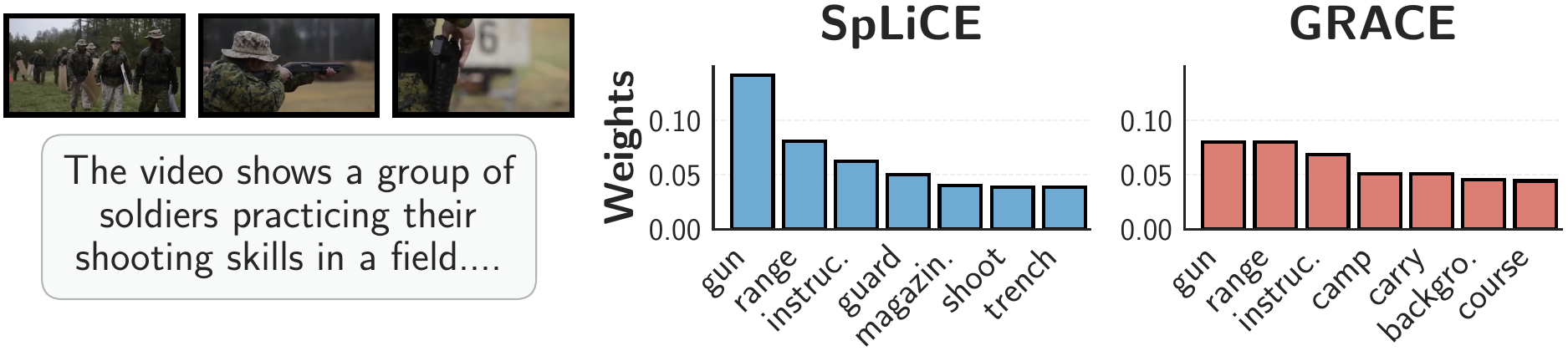}
    \end{subfigure}
    \vspace{1em}
    \begin{subfigure}[b]{0.48\linewidth}
        \centering
        \includegraphics[width=\linewidth]{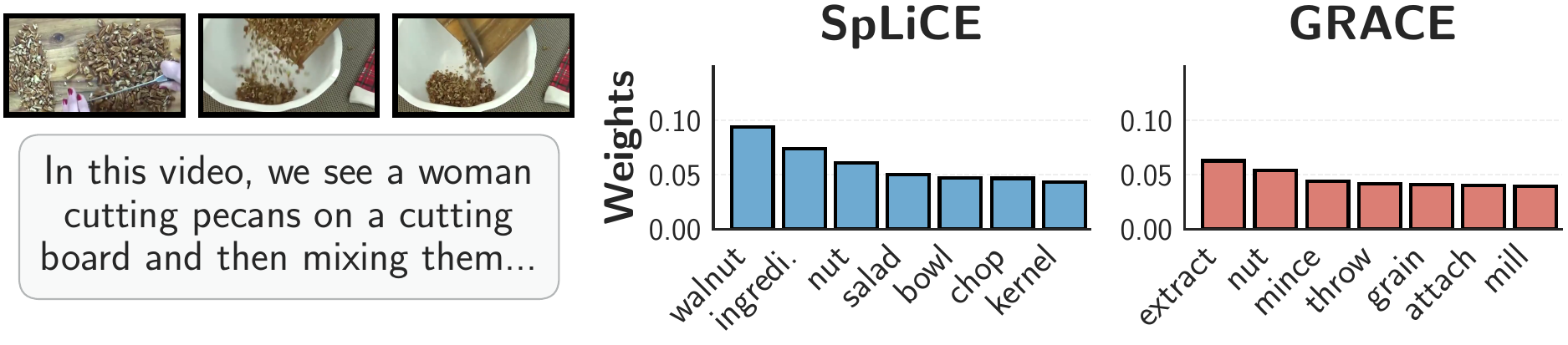}
    \end{subfigure}
    \hfill
    \begin{subfigure}[b]{0.48\linewidth}
        \centering
        \includegraphics[width=\linewidth]{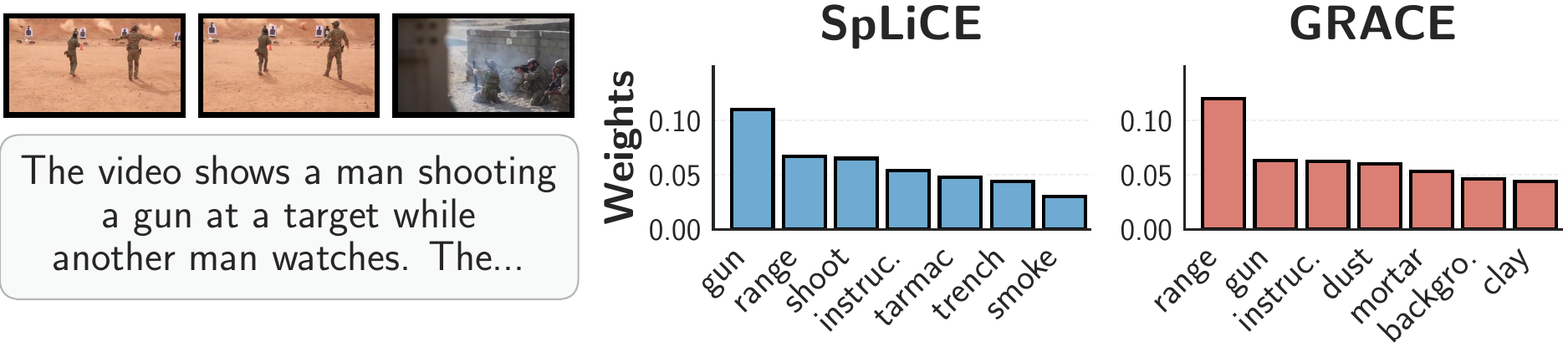}
    \end{subfigure}
    \vspace{1em}
    \begin{subfigure}[b]{0.48\linewidth}
        \centering
        \includegraphics[width=\linewidth]{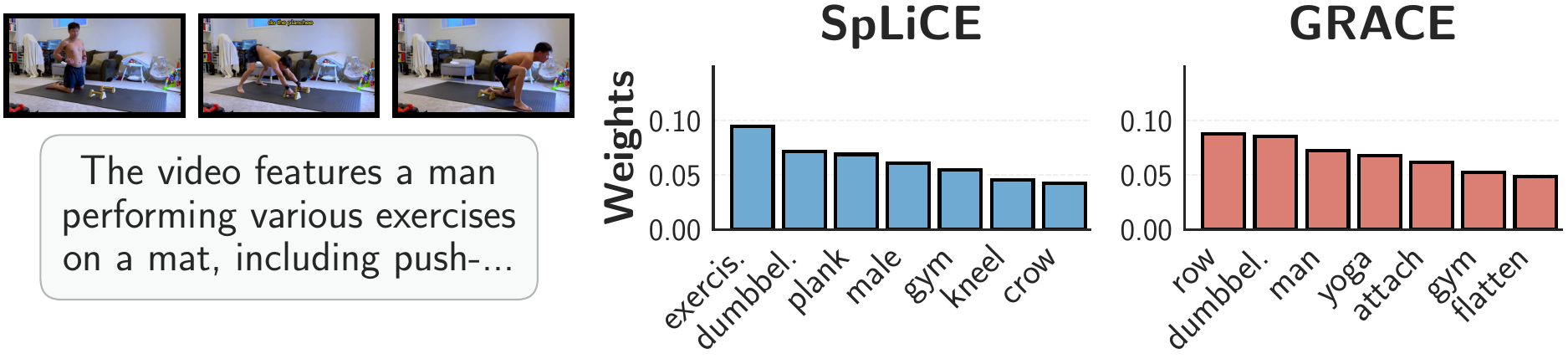}
    \end{subfigure}
    \hfill
    \begin{subfigure}[b]{0.48\linewidth}
        \centering
        \includegraphics[width=\linewidth]{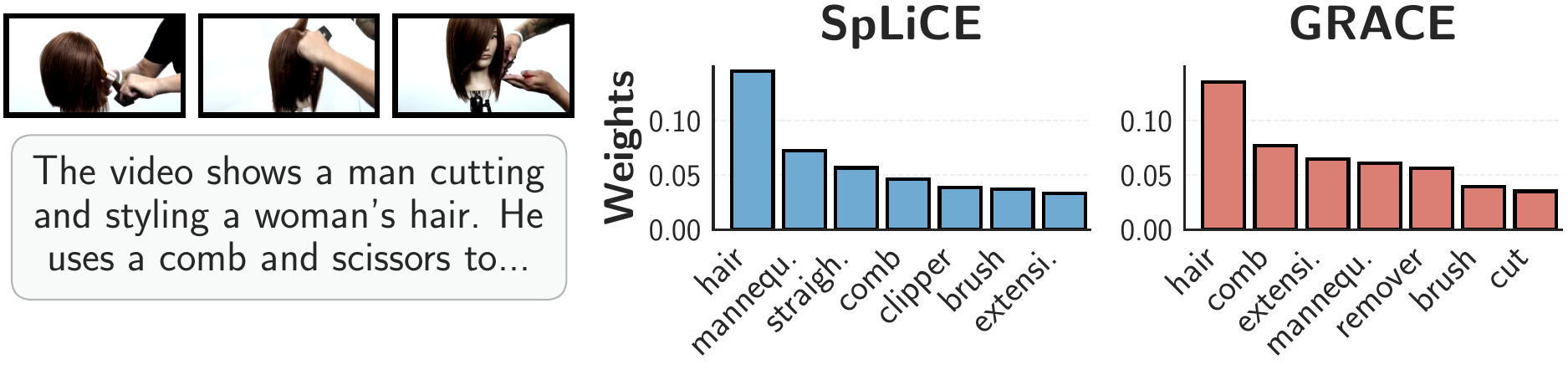}
    \end{subfigure}
    \vspace{1em}
    \begin{subfigure}[b]{0.48\linewidth}
        \centering
        \includegraphics[width=\linewidth]{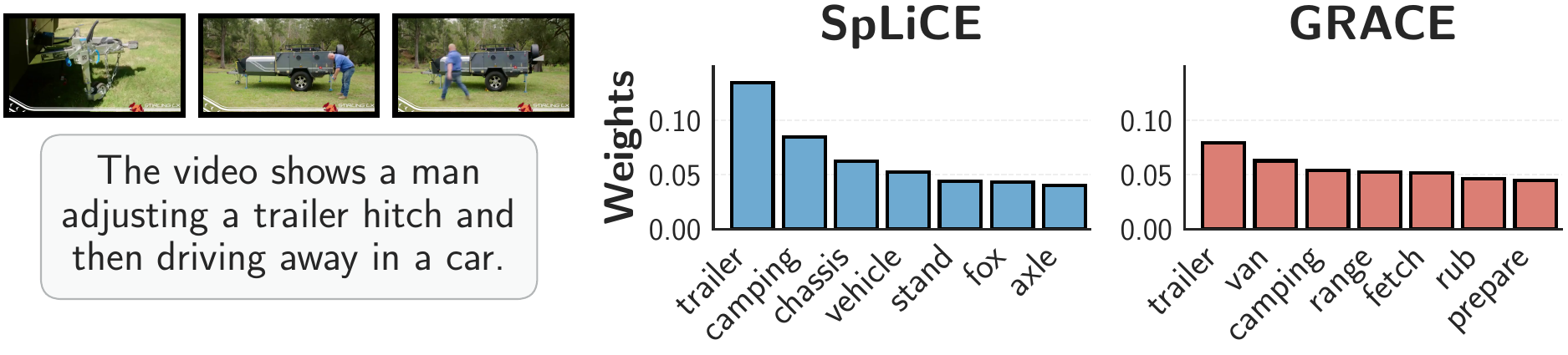}
    \end{subfigure}
    \hfill
    \begin{subfigure}[b]{0.48\linewidth}
        \centering
        \includegraphics[width=\linewidth]{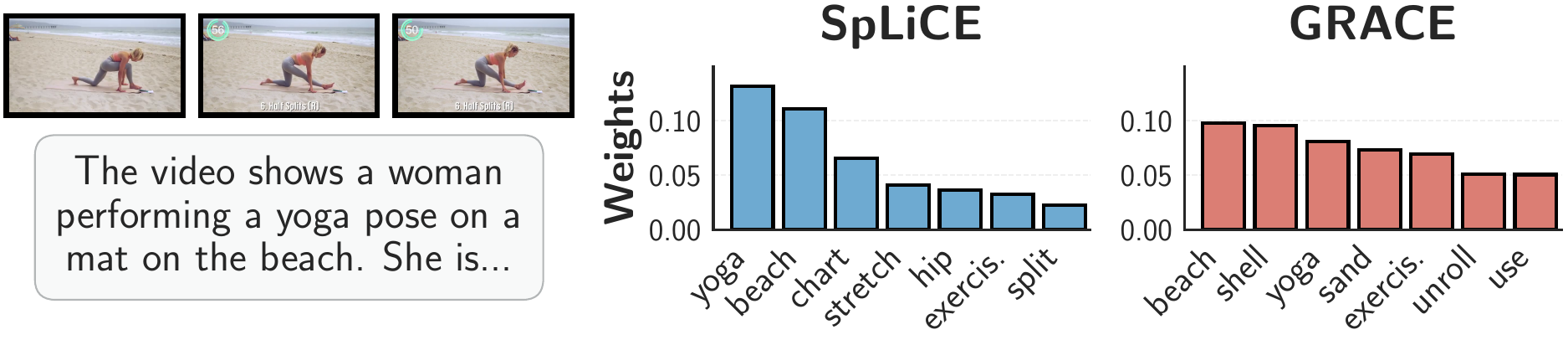}
    \end{subfigure}
    \caption{
        \textbf{Reconstruction weights on ConVisBench~\cite{liberatori2025convisbench}.} 
        For each video, we show a few uniformly sampled frames, the ground truth caption and the Top-k reconstruction weights from the two methods. Both \splice and \ours recover semantically meaningful concepts from the input video, even though \splice tends to allocate more reconstruction mass to the first few concepts.
    }
    \label{fig:appendix_qualitatives_examples}
\end{figure*}

\begin{figure*}[t]
    \centering
    \begin{subfigure}[b]{0.95\linewidth}
        \centering
        \includegraphics[width=\linewidth, trim={0 0 0 0}]{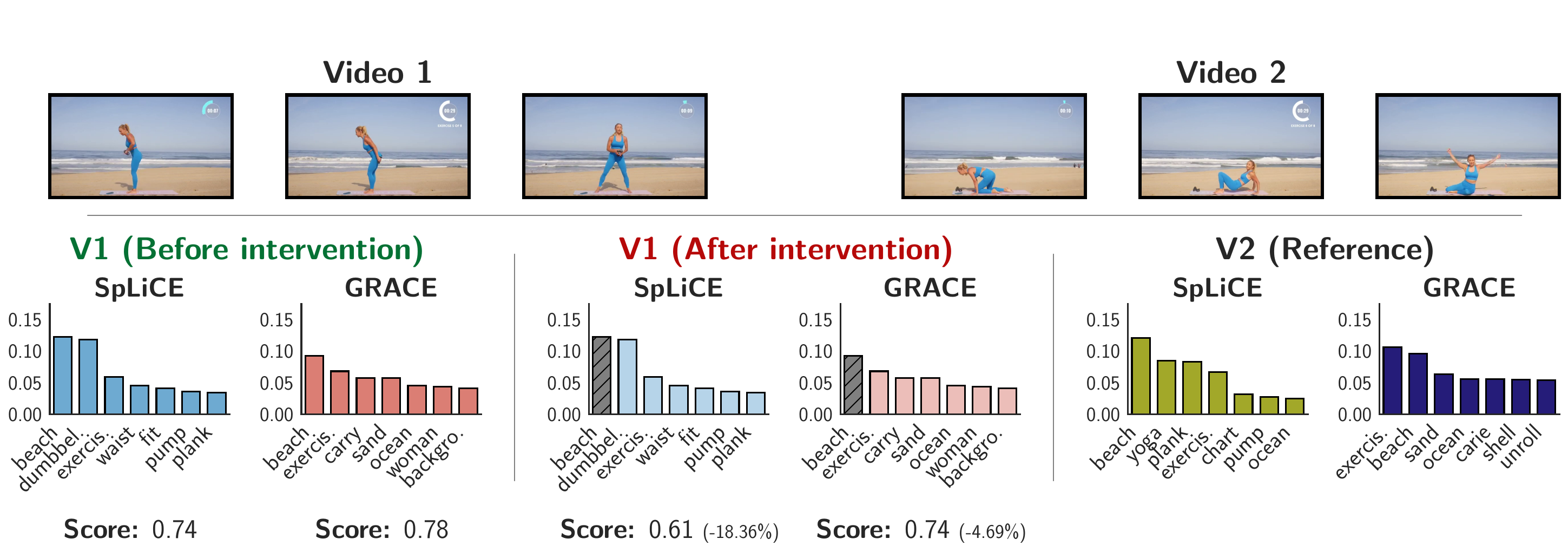}
        \caption{Comparison between videos \texttt{1514\_005} and \texttt{1514\_008}.}
        \label{fig:appendix_interventions__a}
    \end{subfigure}
    \smallskip
    \begin{subfigure}[b]{0.95\linewidth}
        \centering
        \includegraphics[width=\linewidth, trim={0 0 0 0}]{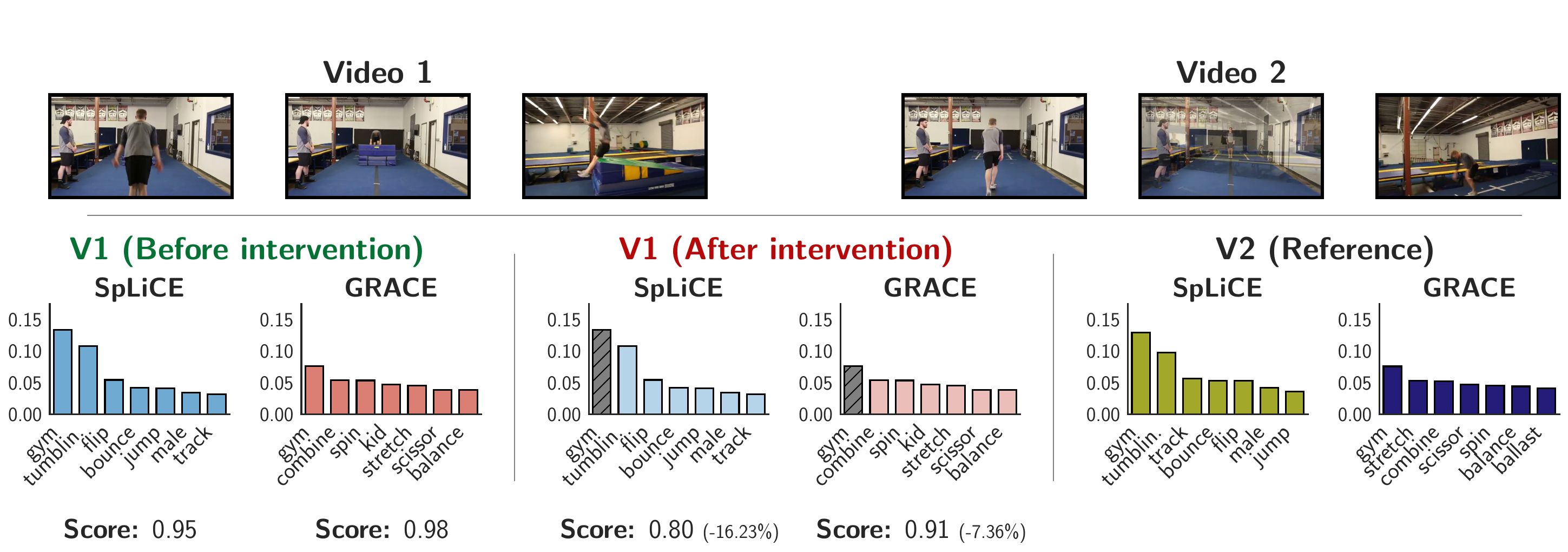}
        \caption{Comparison between videos \texttt{1525\_004} and \texttt{1525\_005}.}
        \label{fig:appendix_interventions__b}
    \end{subfigure}
    \smallskip
    \begin{subfigure}[b]{0.95\linewidth}
        \centering
        \includegraphics[width=\linewidth, trim={0 0 0 0}]{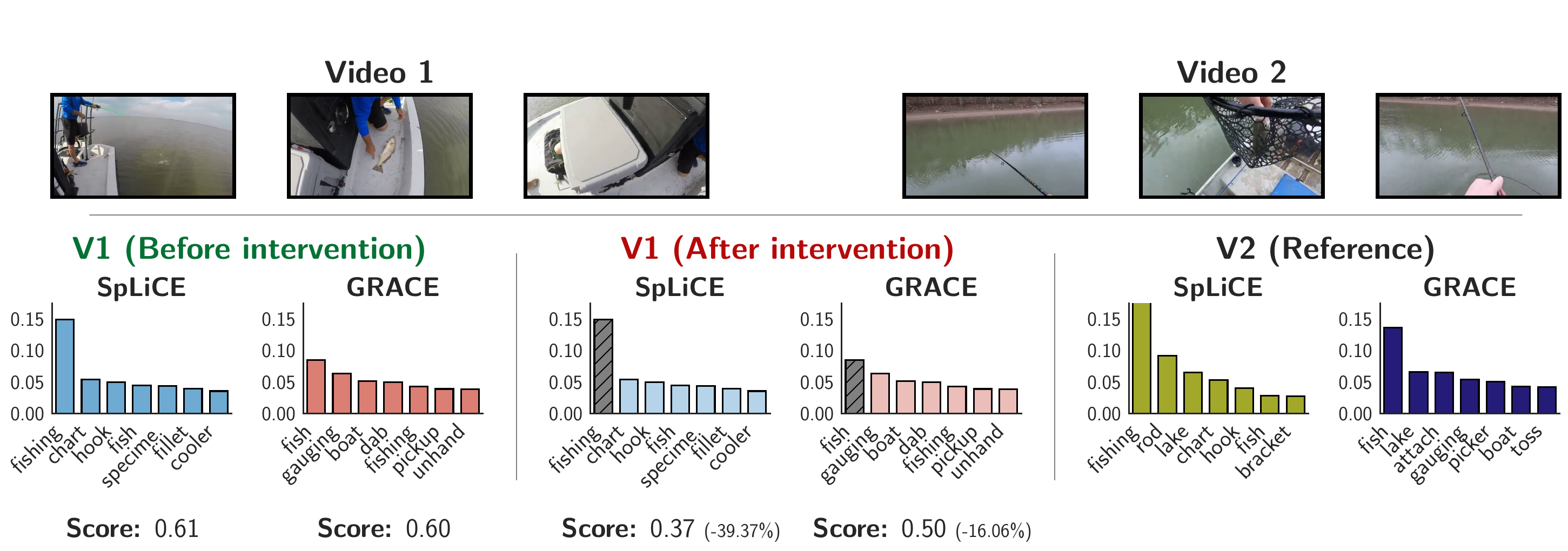}
        \caption{Comparison between videos \texttt{3637\_006} and \texttt{3941\_003}.}
        \label{fig:appendix_interventions__c}
    \end{subfigure}
    \smallskip
    \caption{\textbf{Concept intervention on ConVisBench~\cite{liberatori2025convisbench}.} We artificially zero out the concept corresponding to the highest reconstruction weight to test the ability of these models in capturing all the relevant concepts in the input. After the intervention, the two videos remain similar according to multiple semantic axes. However, the similarity score predicted by \splice drops significantly after the intervention, while \ours shows a more robust behavior.}
    \label{fig:appendix_interventions}
\end{figure*}
We present additional qualitative results, showing the concepts matched in the reconstruction process by \splice and \ours (Figure~\ref{fig:appendix_qualitatives_examples}) and the robustness of our method to concept interventions (Figure~\ref{fig:appendix_interventions}).

Both \splice and \ours recover semantically meaningful concepts from the input video, even though the reconstruction weights from \splice tend to concentrate in the first few concepts.
This behavior indicates that \splice is over-focusing on these concepts, compared to \ours that distributes the reconstruction weights more evenly over the most relevant concepts.
With the intervention, we evaluate how much the model is relying on these concepts to predict the similarity between the two videos. 
Even when removing the most significant concept, the two videos remain similar on multiple semantic axes, \eg, the location or the main subject.
In this scenario, the similarity score predicted by \splice drops significantly, while \ours is more robust to these interventions by distributing the reconstruction weights more evenly across the concepts present in the input.

%% file: appendix/sections/limitations.tex
\section{Limitations and future works}\label{sec:appendix_limitations}
Our approach is a significant effort towards making Vision-Language Models more interpretable, decomposing visual embeddings into human-understandable concepts.
However, the notion of a ``\emph{concept}'' is itself not uniquely defined and may be considered quite subjective. 
Empirically, in our work, we define a concept as a semantic unit that a human would use to describe the same input, while acknowledging that this definition is necessarily approximate and not very rigorous.
Because of this definition, and due to the potential incompleteness of the vocabulary source, some concepts may be absent from the concept vocabulary or may not be readily describable through text~\cite{shang2024incremental}, limiting the coverage and, consequently, the interpretability of the resulting decomposition.
Modeling the residual information that cannot be expressed through concepts is a very relevant future research direction. 
Regarding the interpretability and alignment of the concept decompositions with human perception, we observe a higher correlation with human judgments on ConVisBench compared to existing approaches. This suggests that the concepts extracted by our model are better aligned with the factors considered by human annotators when assessing the similarity between different videos. 
Even though this is a strong indication that the model embeds a concept-based structure, we do not establish a strict causal relation between the concept decomposition produced by \ours and the model behavior on downstream tasks. 
Notably, none of the prior methods reported on ConVisBench establish a strict causal implication between the concept decomposition and the model's behavior; they instead rely on correlation evidence, \ie, alignment with human judgment. Although we do not claim a formal causal relation between the concept decomposition produced by \ours and the model's behavior, our approach improves the understanding of the concept-based structure of VLMs like CLIP.
Similarly, our approach does not model how concepts evolve over time in a video, as we produce clip-level explanations. 
We provide preliminary evidence about the spatial grounding of the concepts obtained from the decomposition. However, our approach does not model temporal grounding of the decomposed concepts. 
We leave these as future work.